\documentclass[runningheads]{llncs}

\usepackage{eccv}

\usepackage{eccvabbrv}

\usepackage{graphicx}
\usepackage{amsmath}
\usepackage{booktabs}
\usepackage{booktabs}
\usepackage{xcolor}
\usepackage{colortbl}
\usepackage{pifont}
\newcommand{\cmark}{\ding{51}}
\usepackage{algorithmicx}
\usepackage{algorithm}
\usepackage{algpseudocode}
\usepackage{amsmath, amssymb}
\usepackage{subcaption}
\usepackage{url}
\usepackage{kotex}
\usepackage{multirow}
\usepackage{pifont}

\usepackage[accsupp]{axessibility}  

\usepackage{hyperref}

\usepackage{orcidlink}

\begin{document}

\title{Residual SODAP: Residual Self-Organizing Domain-Adaptive Prompting with Structural Knowledge Preservation for Continual Learning} 

\titlerunning{Residual SODAP}

\author{
Gyutae Oh\inst{1}\thanks{First author:alswo740012@g.skku.edu} \and
Jungwoo Bae\inst{1} \and
Jitae Shin\inst{1}\thanks{Corresponding author: jtshin@skku.edu}
}

\authorrunning{Gyutae Oh. et al.}

\institute{
Department of Electrical and Computer Engineering,\\
Sungkyunkwan University,\\
Suwon 16419, Republic of Korea\\
\email{alswo740012@g.skku.edu, baejungwoo@skku.edu, jtshin@skku.edu}
}

\maketitle

\begin{abstract}
Continual learning (CL) suffers from catastrophic forgetting, which is exacerbated in domain-incremental learning (DIL) where task identifiers are unavailable and storing past data is infeasible. 
While prompt-based CL (PCL) adapts representations with a frozen backbone, we observe that prompt-only improvements are often insufficient due to suboptimal prompt selection and classifier-level instability under domain shifts. 
We propose Residual SODAP, which jointly performs prompt-based representation adaptation and classifier-level knowledge preservation. 
Our framework combines $\alpha$-entmax sparse prompt selection with residual aggregation, data-free distillation with pseudo-feature replay, prompt-usage--based drift detection, and uncertainty-aware multi-loss balancing. 
Across three DIL benchmarks without task IDs or extra data storage, Residual SODAP achieves state-of-the-art AvgACC/AvgF of 0.850/0.047 (DR), 0.760/0.031 (Skin Cancer), and 0.995/0.003 (CORe50).

\end{abstract}

\section{Introduction}
\label{sec:intro}
Artificial intelligence has progressively evolved toward emulating the cognitive capabilities of living organisms~\cite{hassabis2017neuroscience,yin2020dreaming}. While biological systems flexibly accommodate new information without compromising previously acquired knowledge, artificial neural networks frequently fail to do so~\cite{parisi2019continual,mccloskey1989catastrophic,tadros2022sleep}—a phenomenon known as catastrophic forgetting. In dynamic environments, the ability to continuously assimilate new knowledge is therefore essential not only for biological agents but also for artificial intelligence systems~\cite{liu2024continual,wang2024toward,hadsell2020embracing}.
In the visual domain, variations in acquisition conditions, devices, and environments give rise to heterogeneous data distributions, causing models trained on one domain to suffer sharp performance degradation in another. Continual learning (CL) has been proposed to address this challenge~\cite{buzzega2020dark,kim2024one,schwarz2018progress,yu2024boosting}. Moreover, real-world deployments often impose strict constraints on data export and long-term storage due to privacy regulations~\cite{smith2023closer}, driving growing demand for CL methods that remain effective under such restrictions~\cite{wu2024continual}.
Against this backdrop, we address Domain Incremental Learning (DIL), where a model must incorporate new domain knowledge without retaining any previous data. Prompt-based continual learning (PCL) has recently emerged as a promising paradigm for this setting~\cite{kim2024one,wang2022dualprompt,wang2022learning}; however, existing PCL methods exhibit two key limitations.

\noindent\textbf{Limited prompt selection schemes.} Hard selection (Top-k) methods~\cite{wang2022learning} restrict the model to a fixed subset of prompts, limiting expressiveness and potentially causing training instability due to the non-differentiability of the selection process~\cite{smith2023coda}. Soft selection (Softmax) methods~\cite{kim2024one} alleviate this issue by enabling gradient-based learning over all prompts; however, the weighted-sum nature of softmax allows even irrelevant prompts to exert influence, leading to noise accumulation~\cite{wang2022kvt}.

\noindent\textbf{Neglect of the classifier structure.} Existing PCL methods have predominantly focused on prompt or prompt-pool design to improve CL performance and mitigate forgetting~\cite{kim2024one,wang2022learning,wang2022dualprompt}. While this enables effective representation adaptation, \cite{liu2020more} suggests that classifier design can be equally important. Through simple empirical observations, we show that jointly considering the classifier structure alongside prompt design is indeed necessary. We note that, although multi-head classifiers have been proposed to mitigate forgetting by maintaining separate heads per task~\cite{wang2024comprehensive}, such methods generally assume task identity (Task-ID) at inference time. Since our DIL setting does not provide explicit Task-ID, our approach differs from these existing multi-head methods in the applicable problem scope.
Building on these observations, we propose \textbf{Residual Self-Organizing Domain-Adaptive Prompting (Residual SODAP)}, which jointly improves both the prompt pool and the classifier. \textbf{Our contributions are fourfold}:
\begin{itemize} 
\item \textbf{$\alpha$-Entmax-based Residual Prompt Selection.} We enhance the input query with a memory bank and perform sparse prompt selection via $\alpha$-entmax~\cite{peters2019sparse} combined with residual connections, structurally preserving prior knowledge through frozen prompts while adapting to new domains. 
\item \textbf{Statistical Knowledge Preservation.} Without accessing past data, we store class-wise feature distribution statistics and leverage them for feature distillation and pseudo feature replay during classifier training, thereby mitigating catastrophic forgetting. 
\item \textbf{Prompt Usage-based Drift Detection.} We detect domain drift by jointly monitoring entropy changes in prompt selection patterns and structural shifts in selection behavior, enabling automatic prompt expansion and switching. 
\item \textbf{Uncertainty Weighting.} By learning the uncertainty of each loss term, we enable stable joint optimization without manual weight tuning. \end{itemize}

\section{Related Work}
\label{sec:Related Work}

Domain adaptation (DA)~\cite{wang2018deep} and domain generalization (DG)~\cite{wang2022generalizing} also address distribution shifts across domains~\cite{li2024comprehensive,sun2020test}. However, DA typically assumes prior or concurrent access to target-domain data, whereas DG is usually trained offline on multiple source domains without observing the target domain. As a result, these paradigms rely on different assumptions than our sequential setting where domains arrive over time. In contrast, this study focuses on a DIL setting where domains are presented sequentially, past data is inaccessible, and no Task-ID is provided~\cite{wang2024comprehensive}. Due to these fundamental differences in problem definition and evaluation protocols, we restrict our comparisons to continual learning baselines.

\noindent\textbf{Continual Learning.}
Traditional CL methods can be broadly categorized into three groups~\cite{wu2024continual}. First, Regularization-based CL (Reg-CL) mitigate catastrophic forgetting by constraining changes to parameters deemed important during training. However, under complex distribution shifts or challenging settings, performance may degrade due to limited expressivity and adaptability~\cite{schwarz2018progress,zenke2017continual,li2017learning}. Second, Rehearsal-based CL (Reh-CL) achieve strong performance by storing or replaying a subset of samples from previous tasks and training them jointly with current data, but their applicability is limited in practice due to constraints on data storage and reuse~\cite{buzzega2020dark,yin2020dreaming,lopez2017gradient}. Third, Architectural-based CL (Arch-CL) alleviate forgetting by progressively expanding the model architecture, yet they often require a substantial number of additional parameters and increased computational cost~\cite{he2025cl,yu2024boosting,rusu2016progressive}.

\noindent\textbf{Prompt-based Continual Learning.}
Recently, PCL has attracted attention because it can achieve high performance by training only a small number of prompts or a prompt pool, without requiring past-data storage or additional Task-ID~\cite{oh2025multi,wang2024comprehensive}. Owing to these properties, a range of studies has explored PCL from both performance and efficiency perspectives. Nevertheless, existing methods often exhibit structural limitations in prompt selection. Hard selection (Top-k) methods~\cite{wang2022learning} use only a subset of selected prompts, limiting expressiveness, and may suffer from training instability because gradients are not directly propagated through the selection process~\cite{smith2023coda}. In contrast, soft selection (Softmax) methods~\cite{kim2024one} can utilize all prompts and learn them via gradients; however, due to the weighted-sum nature of softmax, even irrelevant prompts can exert influence, leading to accumulated noise~\cite{wang2022kvt}.

\begin{figure*}[ht!]
\centering
\includegraphics[width=0.8\textwidth]{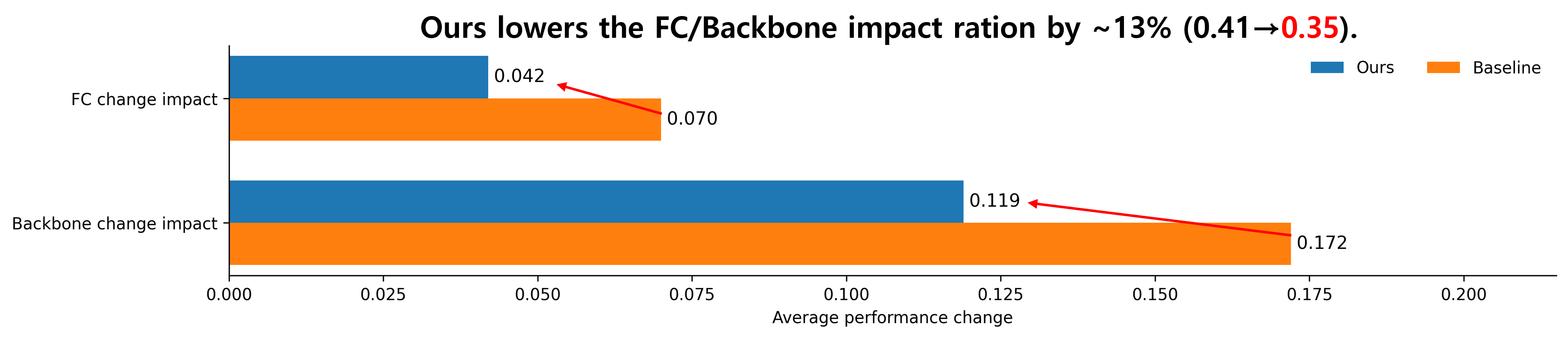}
\caption{PCL exhibits classifier-level forgetting. We apply the cross-composition (backbone $\times$ classifier) diagnostic of \cite{liu2020more} to a \cite{kim2024one}-based baseline under the skin-cancer DIL setting. Residual SODAP mitigates performance degradation caused by classifier instability and improves overall accuracy. The experimental setup and interpretation are provided in Sup.~\ref{supp:A}--\ref{supp:B}.}
\label{figure1}
\end{figure*}

\noindent\textbf{Distinctiveness and Proposed Approach.}
While our method also involves limited parameter expansion and prompt selection, it does not rely on naive architectural expansion. Instead, we focus on encouraging harmony between prior knowledge and new knowledge by leveraging memory-bank-based query enhancement and feature distribution statistics. This design is fundamentally different from existing Arch-CL methods in both design philosophy and objectives. Moreover, to mitigate the noise issue inherent to softmax-based selection while still utilizing the full prompt pool, we introduce sparse prompt selection based on $\alpha$-entmax~\cite{peters2019sparse}. Furthermore, we propose a mechanism that dynamically reallocates prompts according to domain shifts by tracking changes in prompt selection patterns.

\noindent\textbf{Classifier Instability under Domain Shifts.}
PCL has shown that prompt-structure design plays a critical role in mitigating catastrophic forgetting by enabling effective representation adaptation. However, inspired by the analysis of \cite{liu2020more}, we empirically observe that prompt-only improvements are not sufficient: while the backbone representation is well preserved through prompt-based adaptation, a clear degradation still emerges at the classifier level as domain-incremental training proceeds (Fig.~\ref{figure1}). This indicates that forgetting in PCL is driven not only by representation drift but also by decision-boundary instability in the classifier. Motivated by this, we propose Residual SODAP, a framework that simultaneously achieves prompt-based representation adaptation and classifier-level knowledge preservation. As shown in Fig.~\ref{figure1}, our method consistently improves the classifier-level behavior over the baseline and yields a more favorable accuracy--forgetting trade-off across domains, demonstrating stronger overall robustness under domain-incremental training. The detailed experimental setup and full cross-composition analysis are provided in Sup.~\ref{supp:B}.

\begin{figure*}[ht!]
\centering
\includegraphics[width=0.8\textwidth]{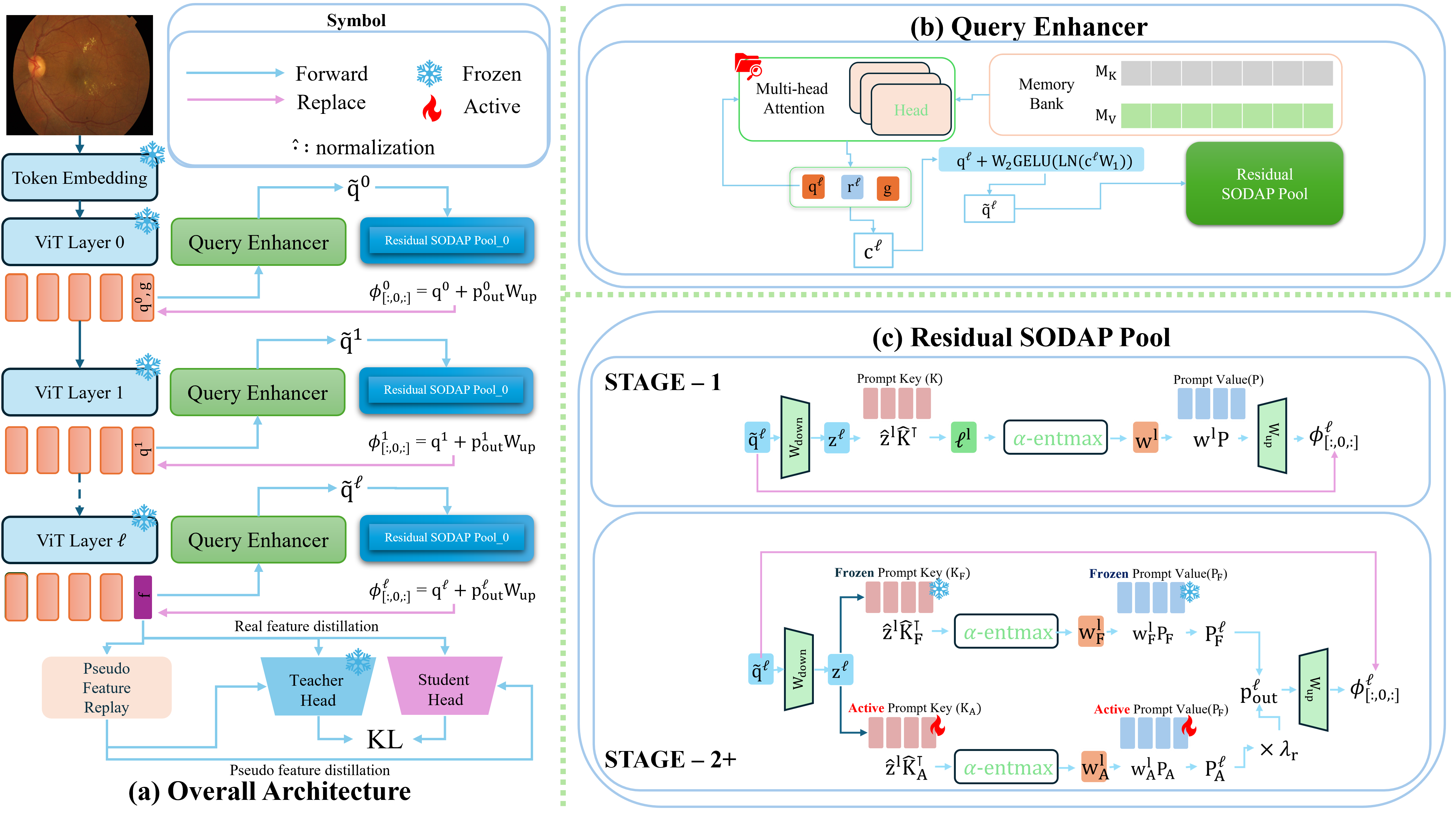}
\caption{Overview of Residual SODAP.}
\label{figure2}
\end{figure*}

\section{Proposed Method}
\label{sec:method}
This section presents the core components of Residual SODAP in four parts:
(i)~$\alpha$-Entmax-based residual prompt selection,
(ii)~Statistical Knowledge Preservation via Statistics-based Pseudo Replay,
(iii)~prompt-usage-pattern-based drift detection (PUDD), 
(iv)~Uncertainty Weighting. The overall architecture is illustrated in Fig.~\ref{figure2}.

\subsection{$\alpha$-Entmax-based residual prompt selection}
\label{subsec:prompt_selection}

\subsubsection{3.1.1 Input and query enhancement}
\label{sssec:query_augment}

At each Transformer layer~$l$, we denote the token representations as $\mathbf{\Phi}^{(l)}\in\mathbb{R}^{B\times(1+S)\times D}$, and use the classification token (CLS) representation
$\mathbf{q}^{(l)}=\mathbf{\Phi}^{(l)}_{[:,0,:]}\in\mathbb{R}^{B\times D}$ as the query.
Here, $S$ is the number of patches (or input tokens), and index~$0$ denotes the CLS token.
Since the CLS token contains an aggregated representation of the input, it is suitable as a query reflecting the semantic characteristics of each sample, and
$B$ denotes the batch size and $D$ denotes the hidden dimension.

We further define the initial CLS token right after embedding,
$\mathbf{g}=\mathbf{\Phi}^{(0)}_{[:,0,:]}\in\mathbb{R}^{B\times D}$,
as a global context shared across all layers.
This enables deeper layers to still reference the original input characteristics and mitigates the locality issue that arises when relying only on the current layer CLS.
The query enhancer integrates the current-layer query~$\mathbf{q}^{(l)}$, the global context~$\mathbf{g}$,
and the retrieved signal~$\mathbf{r}^{(l)}$ from the memory bank.

Memory retrieval is performed via Multi-Head Attention (MHA)~\cite{vaswani2017attention} over a learnable memory bank,
and the memory bank is shared across all layers, providing globally accumulated information throughout training as a global context across layers.
The memory bank consists of learnable key--value pairs
$(\mathbf{M}_{K},\mathbf{M}_{V})\in\mathbb{R}^{M\times D}$, where $M$ is the number of memory size.
The overall process is as follows:
\begin{align}
  \mathbf{r}^{(l)}
    &= \mathrm{MHA}\!\bigl(\mathbf{q}^{(l)},\;\mathbf{M}_{K},\;\mathbf{M}_{V}\bigr)
       \in\mathbb{R}^{B\times D}, \label{eq:mem_read}\\
  \mathbf{c}^{(l)}
    &= \bigl[\mathbf{q}^{(l)};\;\mathbf{g};\;\mathbf{r}^{(l)}\bigr]
       \in\mathbb{R}^{B\times 3D}, \label{eq:concat}\\
  \tilde{\mathbf{q}}^{(l)}
    &= \mathbf{q}^{(l)}
       + \mathbf{W}_{2}\,\mathrm{GELU}\!\bigl(\mathrm{LN}(\mathbf{c}^{(l)}\mathbf{W}_{1})\bigr)
       \in\mathbb{R}^{B\times D}, \label{eq:query_aug}
\end{align}
where $[\,\cdot\,;\,\cdot\,;\,\cdot\,]$ denotes concatenation along the feature dimension,
$\mathrm{LN}(\cdot)$ denotes LayerNorm, and
$\mathbf{W}_{1}\in\mathbb{R}^{3D\times d_{a}}$ and
$\mathbf{W}_{2}\in\mathbb{R}^{d_{a}\times D}$ form a 2-layer adapter with bottleneck dimension~$d_{a}$.
The residual connection prevents loss of the original query information and helps stable optimization.
The memory bank $(\mathbf{M}_{K},\mathbf{M}_{V})$ performs the write operation using a gradient-free EMA~\cite{tarvainen2017mean} to accumulate new information gradually without abrupt changes while maintaining training stability of the read path.

\subsubsection{3.1.2 Bottleneck compression and sparse selection via $\alpha$-Entmax}
\label{sssec:sparse_selection}
To mitigate catastrophic forgetting while preserving useful prior prompts, we perform input-conditioned prompt routing using the enhanced query from Sec.~\ref{sssec:query_augment}. 
Given the enhanced query $\tilde{\mathbf{q}}^{(l)}\in\mathbb{R}^{B\times D}$ at Transformer layer $l$, we project it into the prompt bottleneck space as $\mathbf{z}^{(l)}=\tilde{\mathbf{q}}^{(l)}\mathbf{W}_{\mathrm{down}}\in\mathbb{R}^{B\times d_a}$ with $\mathbf{W}_{\mathrm{down}}\in\mathbb{R}^{D\times d_a}$, where $d_a$ denotes the bottleneck dimension. 
The prompt pool is represented as key--value pairs $(\mathbf{P}_K,\mathbf{P}_V)\in\mathbb{R}^{N\times d_a}$, where $N$ is the prompt-pool size.

We compute prompt-selection logits by cosine similarity, $\boldsymbol{\ell}^{(l)}=\hat{\mathbf{z}}^{(l)}\hat{\mathbf{P}}_K^\top\in\mathbb{R}^{B\times N}$, where $\hat{(\cdot)}$ denotes row-wise $\ell_2$ normalization. 
To suppress unnecessary prompt activations, we apply $\alpha$-entmax~\cite{peters2019sparse} instead of softmax, which yields sparse normalized weights and can assign exact zeros to low-scoring prompts for $\alpha>1$. 
For a logit vector $\boldsymbol{\ell}\in\mathbb{R}^{N}$ (applied row-wise to $\boldsymbol{\ell}^{(l)}$), the $j$-th component of $\alpha$-entmax is defined as
\begin{equation}\label{eq:entmax}
\bigl[\alpha\text{-}\mathrm{entmax}(\boldsymbol{\ell})\bigr]_j
=
\left[
\frac{\alpha-1}{\alpha}\bigl(\ell_j-\tau(\boldsymbol{\ell})\bigr)
\right]_+^{\frac{1}{\alpha-1}},
\end{equation}
where $j$ indexes the prompt dimension, $[x]_+=\max(x,0)$, and $\tau(\boldsymbol{\ell})$ is chosen such that $\sum_{j=1}^{N}w_j=1$ for $\mathbf{w}=\alpha\text{-}\mathrm{entmax}(\boldsymbol{\ell})$. 
The final input-conditioned prompt representation is obtained as $\mathbf{p}^{(l)}=\alpha\text{-}\mathrm{entmax}\!\bigl(\boldsymbol{\ell}^{(l)}\bigr)\mathbf{P}_V$. 
In this work, we use $\alpha=1.5$, which provides a practical trade-off between softmax ($\alpha\to1$) and sparsemax ($\alpha=2$)~\cite{peters2019sparse}.


\subsubsection{3.1.3 Residual combination and output injection}
\label{sssec:residual_injection}
Residual SODAP mitigates catastrophic forgetting by partitioning the prompt pool into a frozen set $\mathcal{F}$ and an active set $\mathcal{A}$, and expanding the pool when necessary.
In Stage~1, all prompts are trainable; thus, we simply route over the full pool.
From Stage~2 onward, we form $\mathcal{F}$ and $\mathcal{A}$ based on the prompt-pool state from the previous stage and determine whether to expand the pool, as described in Sec.~\ref{subsec:pudd}.
In implementation, the prompt pool is maintained as full key--value tensors $(\mathbf{P}_K,\mathbf{P}_V)\in\mathbb{R}^{N\times d_a}$, while $\mathcal{F}$ and $\mathcal{A}$ are managed as index sets.
During training, gradients are blocked for the frozen prompts in $\mathcal{F}$.
From Stage~2 onward, we perform sparse routing independently on $\mathcal{F}$ and $\mathcal{A}$.
For $\mathcal{G}\in\{\mathcal{F},\mathcal{A}\}$, let $(\mathbf{P}_{K,\mathcal{G}},\mathbf{P}_{V,\mathcal{G}})\in\mathbb{R}^{|\mathcal{G}|\times d_a}$ denote the corresponding key--value sub-tensors.
We compute subset logits by cosine similarity,
$\boldsymbol{\ell}_{\mathcal{G}}^{(l)}=\hat{\mathbf{z}}^{(l)}\hat{\mathbf{P}}_{K,\mathcal{G}}^\top\in\mathbb{R}^{B\times|\mathcal{G}|}$,
and obtain the corresponding prompt representation via $\alpha$-entmax routing:
\begin{equation}\label{eq:subset_route}
\mathbf{p}_{\mathcal{G}}^{(l)}
=
\alpha\text{-}\mathrm{entmax}\!\bigl(\boldsymbol{\ell}_{\mathcal{G}}^{(l)}\bigr)\mathbf{P}_{V,\mathcal{G}},
\qquad \mathcal{G}\in\{\mathcal{F},\mathcal{A}\}.
\end{equation}
Here, $\alpha$-entmax is applied row-wise over the subset prompt dimension.

We then combine the frozen and active prompts in a residual form:
\begin{equation}\label{eq:residual_combine}
\mathbf{p}_{\mathrm{out}}^{(l)}
=
\mathbf{p}_{\mathcal{F}}^{(l)}+\lambda_r\,\mathbf{p}_{\mathcal{A}}^{(l)},
\qquad \lambda_r=0.1,
\end{equation}
and inject the resulting prompt into the CLS token:
\begin{equation}\label{eq:cls_inject}
\mathbf{\Phi}^{(l)}_{[:,0,:]}
=
\mathbf{q}^{(l)}+\mathbf{p}_{\mathrm{out}}^{(l)}\mathbf{W}_{\mathrm{up}},
\qquad
\mathbf{W}_{\mathrm{up}}\in\mathbb{R}^{d_a\times D}.
\end{equation}
This design treats the frozen prompts $\mathcal{F}$ as a stable base to preserve prior knowledge, while constraining the active prompts $\mathcal{A}$ to act as a residual for adaptation, enabling stable learning under sequential domain shifts.


\subsubsection{3.1.4 Auxiliary Loss}
\label{sssec:aux_loss}
We propose two auxiliary losses to improve diversity and stability in prompt pool learning.
In Stage~1, since all prompts are trainable, the losses are applied to the entire pool; from Stage~2 onward, they are applied only to the active set $\mathcal{A}$ to avoid interfering with the frozen set $\mathcal{F}$.
Accordingly, the application set is defined as $\mathcal{S}=\{1,\dots,N\}$ in Stage~1 and $\mathcal{S}=\mathcal{A}$ from Stage~2 onward.

\noindent\textbf{(i) Diversity loss.}
Let $w_{b,i}$ denote the prompt-routing weight assigned to prompt $i$ for sample $b$ (from $\alpha$-entmax routing), and define the batch-averaged usage as
$u_i=\frac{1}{B}\sum_{b=1}^{B} w_{b,i}$.
We encourage diverse prompt values by penalizing similarity between frequently co-activated prompts.
For the normalized prompt value $\hat{\mathbf{P}}_{V,i}=\mathbf{P}_{V,i}/\|\mathbf{P}_{V,i}\|_2$, we define
\begin{equation}\label{eq:div_loss}
\mathcal{L}_{\mathrm{div}}
=
\frac{\sum_{i\neq j\in\mathcal{S}} u_i\,u_j\,\bigl|\langle \hat{\mathbf{P}}_{V,i},\hat{\mathbf{P}}_{V,j}\rangle\bigr|}
{\sum_{i\neq j\in\mathcal{S}} u_i\,u_j}.
\end{equation}
This reduces redundancy by discouraging high similarity among prompt values that are frequently selected together.

\noindent\textbf{(ii) Norm regularization.}
From Stage~2 onward, we constrain the magnitude of \emph{active prompt values} so that they operate only as residuals:
\begin{equation}\label{eq:norm_loss}
\mathcal{L}_{\mathrm{norm}}
=
\frac{1}{|\mathcal{A}|}\sum_{i\in\mathcal{A}}\|\mathbf{P}_{V,i}\|_{2}.
\end{equation}

\subsection{Statistical Knowledge Preservation via Statistics-based Pseudo Replay}
\label{subsec:skp}

In DIL, the data domain shifts sequentially as stages progress, while the class space~$C$ remains fixed across all stages.
Under a rehearsal-free setting, direct access to data from previous stages is unavailable, making it crucial to preserve prior knowledge using stored statistical information.
Instead of replaying inputs directly, we introduce statistical knowledge preservation, maintains the previous decision boundaries via statistics-based pseudo replay at the feature level.

\subsubsection{3.2.1 Saving Knowledge Assets at Stage Transitions}
\label{sssec:knowledge_asset}

After completing training at each stage, we store two knowledge assets to preserve prior knowledge in the next stage.
First, we duplicate and freeze the current-stage classifier head~$h_{t}$ and save it as a teacher head.
Second, we collect class-wise feature statistics over the entire training data of the current stage.
Here, the feature~$\mathbf{f}\in\mathbb{R}^{D}$ is defined as the CLS token of the last Transformer layer,
$\mathbf{f}=\mathbf{\Phi}^{(L)}_{[:,0,:]}$,
and we estimate the class mean $\boldsymbol{\mu}_{c}\in\mathbb{R}^{D}$ and diagonal variance $\boldsymbol{\sigma}_{c}^{2}\in\mathbb{R}^{D}$
using Welford's online algorithm~\cite{welford1962note}.
Specifically, whenever a new sample~$\mathbf{f}$ belonging to class~$c$ is observed, we update the running statistics as:
\begin{align}
n_c &\leftarrow n_c + 1,
&\boldsymbol{\delta} &\leftarrow \mathbf{f} - \boldsymbol{\mu}_c, \label{eq:welford_update1}\\
\boldsymbol{\mu}_c &\leftarrow \boldsymbol{\mu}_c + \boldsymbol{\delta}/n_c,
&\mathbf{Q}_c &\leftarrow \mathbf{Q}_c + \boldsymbol{\delta}\odot(\mathbf{f}-\boldsymbol{\mu}_c). \label{eq:welford_update2}
\end{align}
where $n_{c}$ is the accumulated number of samples,
$\mathbf{Q}_c\in\mathbb{R}^{D}$ is the sum of squared deviations, and
the final variance is computed using the unbiased estimator
$\boldsymbol{\sigma}_{c}^{2}=\mathbf{Q}_c/(n_{c}-1)$.
This approach enables single-pass estimation of statistics without storing the full dataset, providing strong memory efficiency and numerical stability.
In multi-stage training, we maintain cumulative statistics by merging statistics from the previous and current stages using Welford's formulas.

\subsubsection{3.2.2 Knowledge Distillation During the Next Stage Training}
\label{sssec:kd}

When training begins for the next stage, we load the saved classifier head $h_t$ and the class-wise feature statistics, freeze $h_t$ as a teacher head, and initialize a trainable student head $h_s$ by copying the teacher parameters.
To preserve prior knowledge without accessing past data, we compute two distillation losses at each iteration, both formulated as KL divergence~\cite{kullback1951information,hinton2015distilling} with the teacher distribution as the target.

\noindent\textbf{(i) Real feature distillation.}
We pass the real features of the current batch,
$\mathbf{f}_{\mathrm{real}}\in\mathbb{R}^{B\times D}$,
through the teacher and student heads to obtain
$\mathbf{z}_{t}=h_{t}(\mathbf{f}_{\mathrm{real}})\in\mathbb{R}^{B\times C}$ and
$\mathbf{z}_{s}=h_{s}(\mathbf{f}_{\mathrm{real}})\in\mathbb{R}^{B\times C}$,
and align the two output distributions:
\begin{equation}\label{eq:real_kd}
  \mathcal{L}_{\mathrm{real}}
  = \mathrm{KL}\!\Bigl(
      \mathrm{softmax}(\mathbf{z}_{t}/T)
      \;\Big\|\;
      \mathrm{softmax}(\mathbf{z}_{s}/T)
    \Bigr)\cdot T^{2}.
\end{equation}
Here, $T$ is the distillation temperature (default: $2.0$), and the $T^{2}$ factor rescales gradients for stable soft-label distillation~\cite{hinton2014distilling}.

\noindent\textbf{(ii) Pseudo feature replay.}
We replay prior knowledge at the feature level by sampling $K$ pseudo features from the stored class-wise statistics $(\boldsymbol{\mu}_{c},\boldsymbol{\sigma}_{c}^{2})$.
We approximate the feature distribution of class $c$ as a diagonal Gaussian
$\mathcal{N}(\boldsymbol{\mu}_{c},\,\mathrm{diag}(\boldsymbol{\sigma}_{c}^{2}))$
and sample class indices uniformly to mitigate under-representation of minority classes.
Specifically, for $c_{k}\sim\mathrm{Uniform}\{1,\dots,C\}$ and $\boldsymbol{\epsilon}_{k}\sim\mathcal{N}(\mathbf{0},\mathbf{I}_{D})$, we generate pseudo features via the reparameterization trick:
\begin{equation}\label{eq:pseudo_sample}
  \tilde{\mathbf{f}}_{k}
  = \boldsymbol{\mu}_{c_{k}} + \boldsymbol{\epsilon}_{k}\odot\boldsymbol{\sigma}_{c_{k}},
  \qquad
  \boldsymbol{\sigma}_{c_{k}}=\sqrt{\boldsymbol{\sigma}_{c_{k}}^{2}}.
\end{equation}
Equivalently, $\tilde{\mathbf{f}}_{k}\sim\mathcal{N}(\boldsymbol{\mu}_{c_{k}},\,\mathrm{diag}(\boldsymbol{\sigma}_{c_{k}}^{2}))$.
We treat $\tilde{\mathbf{f}}_{k}$ as a constant (stop-gradient) and pass it through the frozen teacher head and the trainable student head.
Let $\tilde{\mathbf{z}}_{t}=h_{t}(\tilde{\mathbf{f}}_{k})\in\mathbb{R}^{C}$ and $\tilde{\mathbf{z}}_{s}=h_{s}(\tilde{\mathbf{f}}_{k})\in\mathbb{R}^{C}$;
then the pseudo replay loss is defined as:
\begin{equation}\label{eq:pseudo_kd}
  \mathcal{L}_{\mathrm{pseudo}}
  = \frac{1}{K}\sum_{k=1}^{K}
    \mathrm{KL}\!\Bigl(
      \mathrm{softmax}\bigl(\tilde{\mathbf{z}}_{t}/T\bigr)
      \;\Big\|\;
      \mathrm{softmax}\bigl(\tilde{\mathbf{z}}_{s}/T\bigr)
    \Bigr)\cdot T^{2}.
\end{equation}
Since $h_t$ is frozen, $\mathcal{L}_{\mathrm{pseudo}}$ backpropagates only through the student head $h_s$.
Here, $K$ is the number of pseudo samples per batch (default: $B$), which enables preserving decision boundaries of previous classes without storing any past data.

Finally, we combine the classification loss $\mathcal{L}_{\mathrm{ce}}$ with
$\mathcal{L}_{\mathrm{real}}$, $\mathcal{L}_{\mathrm{pseudo}}$, $\mathcal{L}_{\mathrm{div}}$, and $\mathcal{L}_{\mathrm{norm}}$.
For clarity, we write the objective as a weighted sum,
\begin{equation}\label{eq:total_loss}
  \mathcal{L}_{\mathrm{total}}
  = \lambda_{\mathrm{ce}}\,\mathcal{L}_{\mathrm{ce}}
    + \lambda_{\mathrm{real}}\,\mathcal{L}_{\mathrm{real}}
    + \lambda_{\mathrm{pseudo}}\,\mathcal{L}_{\mathrm{pseudo}}
    + \lambda_{\mathrm{div}}\,\mathcal{L}_{\mathrm{div}}
    + \lambda_{\mathrm{norm}}\,\mathcal{L}_{\mathrm{norm}}.
\end{equation}
In our implementation, these weights are determined dynamically via the uncertainty weighting scheme in Sec.~\ref{subsec:uw} (Eq.~\ref{eq:uw}).

\subsection{Prompt Usage-based Drift Detection (PUDD)}
\label{subsec:pudd}

In prompt-based continual learning, it is important to promptly detect incoming data from a new domain and to automatically decide responses such as splitting and expanding the prompt pool based on the detection results.
Accordingly, we propose PUDD, which detects domain shifts by monitoring changes in the usage patterns of prompt selection weights.
The core assumption is that when the domain changes, the optimal prompt combination changes as well, leading to significant variations in the prompt selection distribution and the set of used prompts.
In addition, since the usage set defined in PUDD may differ in purpose from the active set~$\mathcal{A}$ during training, we denote it separately as~$\mathcal{S}_{t}$.

\subsubsection{3.3.1 Extracting prompt usage signals}
\label{sssec:usage_signal}

At each iteration~$t$, given the prompt selection weights
$\mathbf{w}_{t}^{(b)}\in\mathbb{R}^{N}$ for samples $b=1,\dots,B$ in the batch, we define the batch average as follows and extract two drift signals from it:
\begin{equation}
  \bar{\mathbf{w}}_{t}
  =\frac{1}{B}\sum_{b=1}^{B}\mathbf{w}_{t}^{(b)}
  \in\mathbb{R}^{N}.
\end{equation}

\noindent\textbf{(i) Selection entropy.}
We measure the uncertainty of the selection distribution by the entropy
$H_{t}=-\sum_{i=1}^{N}\bar{w}_{t,i}\log(\bar{w}_{t,i}+\epsilon)$.
Here, $\epsilon=10^{-10}$ is a numerical stabilization constant to handle cases where $\bar{w}_{t,i}=0$ can occur due to the sparsity of $\alpha$-entmax.
When the domain changes, the fitness of the existing prompt combination decreases and the selection distribution is re-adjusted, which can manifest as increased short-term fluctuations of $H_{t}$.

\noindent\textbf{(ii) Usage set shift.}
We define the usage set as
$\mathcal{S}_{t}=\{i:\bar{w}_{t,i}>\tau_{s}\}$,
where $\tau_{s}$ is a threshold (default: $0.01$) for prompts whose mass exceeds the threshold.
Let the indices of the most recent $W$ iterations form the sliding window
$\mathcal{W}_{t}=\{t-W+1,\dots,t-1\}$,
and define the union of past usage sets within the window as
$\mathcal{S}_{t}^{\mathrm{ref}}=\bigcup_{s\in\mathcal{W}_{t}}\mathcal{S}_{s}$.
We then compute
\begin{equation}\label{eq:iou}
  \mathrm{IoU}_{t}
  = \frac{|\mathcal{S}_{t}\cap\mathcal{S}_{t}^{\mathrm{ref}}|}
         {|\mathcal{S}_{t}\cup\mathcal{S}_{t}^{\mathrm{ref}}|}
\end{equation}
to quantify changes in the usage set.
A lower $\mathrm{IoU}_{t}$ indicates that prompts different from those predominantly used recently are being selected, which we interpret as a signal of domain shift.

\subsubsection{3.3.2 Computing and measuring the drift score}
\label{sssec:drift_score}

We combine the two signals into a single drift score.
Let the moving average and standard deviation of entropy within the sliding window~$\mathcal{W}_{t}$ be
$\bar{H}_{t}=\mathrm{mean}(\{H_{s}\}_{s\in\mathcal{W}_{t}})$ and
$\sigma_{H,t}=\mathrm{std}(\{H_{s}\}_{s\in\mathcal{W}_{t}})$, respectively.
The drift score is then defined as:
\begin{equation}\label{eq:drift_score}
  D_{t}
  = \alpha\cdot\frac{|H_{t}-\bar{H}_{t}|}{\sigma_{H,t}+\epsilon}
  + \beta\cdot\biggl(\frac{1}{\max(\mathrm{IoU}_{t},\,\eta)}-1\biggr)
\end{equation}
where $\alpha=1.0$ and $\beta=0.5$ are the weights of the two signals, and
$\eta$ is a lower-bound clip for IoU (default: $0.1$).
At each iteration, we add $\{H_{t},\,\mathcal{S}_{t}\}$ to the window~$\mathcal{W}_{t}$ and update
$\bar{H}_{t}$, $\sigma_{H,t}$, and $\mathcal{S}_{t}^{\mathrm{ref}}$, using the default window size $W=100$.
After completing training for Stage~$k$, we pass the training data of the next domain~$\mathcal{D}_{k+1}$ through the frozen model in eval mode for one epoch, collecting $D_{t}$ for each batch.
As batches progress, observations from the new domain accumulate in the sliding window and the reference statistics are gradually updated; thus, in early batches, abrupt changes in selection patterns appear as high $D_{t}$ values and then gradually stabilize.
We average the drift scores collected over $L$ adapter layers and $T$ batches as
\begin{equation}\label{eq:drift_mean}
  \bar{D}
  = \frac{1}{LT}\sum_{l=1}^{L}\sum_{t=1}^{T}D_{t}^{(l)}
\end{equation}
and use it as a single scalar representing the strength of the domain shift.

\subsubsection{3.3.3 Drift-proportional pool expansion}
\label{sssec:drift_expansion}

We determine the expansion scale of the prompt pool in proportion to the measured~$\bar{D}$.
Let the current number of active prompts be~$|\mathcal{A}|$ and the expected maximum drift score be~$D_{\max}$.
The number of added prompts is computed as
\begin{equation}\label{eq:expansion}
  E = \mathrm{clamp}\!\left(
    \left\lfloor |\mathcal{A}| \cdot \frac{\bar{D}}{D_{\max}}
    \right\rfloor,\;
    E_{\min},\; E_{\max}
  \right)
\end{equation}
with defaults $D_{\max}=5.0$, $E_{\min}=10$, and $E_{\max}=80$.
$D_{\max}$ is a normalization constant for the drift score and is set independently of the PUDD detection threshold~$\theta$.
This allows adding only a minimal number of prompts under weak domain shifts ($\bar{D}\ll D_{\max}$) to avoid excessive capacity waste, while allocating sufficient learning capacity under strong shifts ($\bar{D}\to D_{\max}$) to ensure adaptability to the new domain.
After expansion, the existing active prompts are moved to the frozen partition~$\mathcal{F}$, and the newly added prompts are assigned to the new active partition~$\mathcal{A}$.

\subsection{Uncertainty Weighting (UW) }
\label{subsec:uw}

Our training objective consists of multiple loss terms, including
$\mathcal{L}_{\mathrm{ce}}$,
$\mathcal{L}_{\mathrm{real}}$,
$\mathcal{L}_{\mathrm{pseudo}}$,
$\mathcal{L}_{\mathrm{div}}$,
and
$\mathcal{L}_{\mathrm{norm}}$.
In this multi-objective setting, manually tuning fixed weights for each loss is not only inefficient, but it is also difficult to maintain an appropriate balance throughout training because the loss scales and their relative importance can change as training progresses.
To address this, we adopt the homoscedastic uncertainty weighting of Kendall et al.~\cite{kendall2018multi} to automatically learn the relative weights among loss terms.

For each loss $\mathcal{L}_{i}$, we introduce a learnable log-variance $s_{i}=\log\sigma_{i}^{2}$ and define the total loss as
\begin{equation}\label{eq:uw}
  \mathcal{L}_{\mathrm{total}}
  = \sum_{i}\bigl(e^{-s_{i}}\,\mathcal{L}_{i} + s_{i}\bigr)
\end{equation}
where losses with higher uncertainty (larger variance) are automatically assigned smaller weights $w_i=e^{-s_i}$, suppressing the influence of noisier signals, while losses with lower uncertainty receive larger weights, encouraging the model to focus on relatively more reliable signals.
At the same time, the regularization term $s_i$ prevents the trivial solution in which $\sigma_i \to \infty$ and all losses are ignored.
As a result, the balance among losses is dynamically adjusted throughout training without additional manual tuning.

In implementation, we initialize all $s_i$ as $s_i^{(0)}=0$ (i.e., $w_i=e^{-s_i}=1$), so that training starts from an equal baseline without prior knowledge.
For numerical stability, we apply an asymmetric clamp to $s_i$ within the range $[-3,\,6]$.
The upper bound ($s_i=6$, $w_i\approx 0.0025$) allows a loss term to be effectively deactivated when necessary, while the lower bound ($s_i=-3$, $w_i\approx 20$) is conservatively set to prevent training instability caused by excessive gradient amplification.
The uncertainty parameters $\{s_i\}$ are optimized jointly with the model parameters using the same optimizer.
\section{Experiments}
\textbf{Experiment Setting.} We aim to mitigate catastrophic forgetting in a DIL setting, where the input domain shifts sequentially over time. Following~\cite{wang2024comprehensive}, we further assume a rehearsal-free constraint, i.e., the original data from past domains cannot be stored. To evaluate this setting, we preprocess a total of seven datasets and construct three experimental scenarios: (i) diabetic retinopathy (DR)~\cite{aptos2019-blindness-detection,LI2019,diabetic-retinopathy-detection}, (ii) skin cancer~\cite{rotemberg2021patient,codella2019skin,kawahara2018seven}, and (iii) CORe50~\cite{lomonaco2017core50} as a general-domain benchmark to assess generalization beyond medical imaging. The dataset preprocessing pipeline, data usage protocol, and all training hyperparameters are detailed in Sup.~\ref{supp:A}. For evaluation, we report AvgACC to measure how well the model maintains average performance throughout DIL training, and AvgF to quantify the extent of knowledge loss on previously seen domains. The definitions and interpretations of these metrics are provided in Sup.~\ref{supp:C}. Unless otherwise stated, all numbers reported in tables are averaged over three independent runs.

\noindent\textbf{Comparison with State-of-the-Art CL Methods.}
We compare Residual SODAP with representative methods from diverse continual learning families, including PCL, Reh-CL, Reg-CL, and Arch-CL, and summarize the results in Table~\ref{Table1}.
Across all benchmarks, our method achieves the strongest or most competitive accuracy, demonstrating robust retention under sequential domain shifts.
In particular, on the DR scenario, Residual SODAP attains the best AvgACC (0.850) and also the lowest AvgF (0.047) among all compared methods, indicating that it improves final performance while stabilizing knowledge preservation over time.
On the Skin Cancer scenario, Residual SODAP again achieves the highest AvgACC (0.760).
While Dual-Prompt yields the lowest forgetting (AvgF 0.012), it suffers from substantially lower AvgACC (0.637), revealing a clear accuracy--forgetting trade-off in this setting.
By contrast, Residual SODAP provides a favorable operating point, delivering the top AvgACC while keeping forgetting comparatively low (AvgF 0.031).
Finally, on the general continual learning benchmark CORe50, our method achieves the best AvgACC (0.995) and the lowest AvgF (0.003), further supporting the generality of our design beyond medical datasets.
Overall, these results show that Residual SODAP consistently improves the accuracy--forgetting balance across heterogeneous CL paradigms and offers a practical DIL solution for medical scenarios with evolving domains.

\begin{table*}[t]
\centering
\scriptsize
\setlength{\tabcolsep}{5pt}
\caption{Performance comparison with continual learning baselines on two medical imaging benchmarks: DR (APTOS~\cite{aptos2019-blindness-detection} $\rightarrow$ DDR~\cite{LI2019} $\rightarrow$ DRD~\cite{diabetic-retinopathy-detection}) and Skin Cancer (ISIC~\cite{rotemberg2021patient} $\rightarrow$ HAM~\cite{codella2019skin} $\rightarrow$ DERM7~\cite{kawahara2018seven}), and a general continual learning benchmark CORe50~\cite{lomonaco2017core50}. Best results are in \textbf{bold}.}
\label{Table1}
\resizebox{1.0\textwidth}{!}{
\begin{tabular}{l l c c c c c c}
\toprule
\textbf{CL-Type [Ref.]} & \textbf{Method}
& \multicolumn{2}{c}{\textbf{DR}}
& \multicolumn{2}{c}{\textbf{Skin Cancer}}
& \multicolumn{2}{c}{\textbf{CORe50}} \\
\cmidrule(lr){3-4}\cmidrule(lr){5-6}\cmidrule(lr){7-8}
& & AvgACC$\uparrow$ & AvgF$\downarrow$ & AvgACC$\uparrow$ & AvgF$\downarrow$ & AvgACC$\uparrow$ & AvgF$\downarrow$ \\
\midrule
PCL~\cite{kim2024one} & OS-Prompt            
& 0.666$\pm$0.062 & 0.132$\pm$0.003 
& 0.682$\pm$0.041 & 0.135$\pm$0.075 
& 0.982$\pm$0.004 & 0.012$\pm$0.012 \\

PCL~\cite{kim2024one} & OS-Prompt++          
& 0.769$\pm$0.027 & 0.113$\pm$0.006 
& 0.725$\pm$0.032 & 0.063$\pm$0.013 
& 0.983$\pm$0.001 & 0.014$\pm$0.002 \\

PCL~\cite{smith2023coda} & Coda-Prompt       
& 0.688$\pm$0.079 & 0.140$\pm$0.044 
& 0.713$\pm$0.040 & 0.041$\pm$0.013 
& 0.974$\pm$0.041 & 0.056$\pm$0.011 \\

PCL~\cite{wang2022dualprompt} & Dual-prompt   
& 0.467$\pm$0.159 & 0.291$\pm$0.065 
& 0.637$\pm$0.065 & \textbf{0.012}$\pm$0.039 
& 0.975$\pm$0.002 & 0.036$\pm$0.001 \\

Arch-CL~\cite{yu2024boosting} & MoE-Adapters  
& 0.716$\pm$0.141 & 0.080$\pm$0.092 
& 0.597$\pm$0.165 & 0.040$\pm$0.016 
& 0.949$\pm$0.041 & 0.043$\pm$0.023 \\

Reh-CL~\cite{buzzega2020dark} & DER++         
& 0.607$\pm$0.121 & 0.288$\pm$0.160 
& 0.722$\pm$0.057 & 0.099$\pm$0.023 
& 0.994$\pm$0.004 & 0.061$\pm$0.009 \\

Reg-CL~\cite{schwarz2018progress} & Online EWC 
& 0.715$\pm$0.151 & 0.174$\pm$0.002 
& 0.708$\pm$0.054 & 0.157$\pm$0.002 
& 0.989$\pm$0.004 & 0.029$\pm$0.001 \\

\midrule
\textbf{Ours} & \textbf{Residual SODAP} 
& \textbf{0.850$\pm$0.001} & \textbf{0.047$\pm$0.003} 
& \textbf{0.760$\pm$0.004} & 0.031$\pm$0.009 
& \textbf{0.995$\pm$0.001} & \textbf{0.003$\pm$0.001} \\

\bottomrule
\end{tabular}
}
\end{table*}

\begin{figure*}[ht!]
\centering
\includegraphics[width=0.6\textwidth]{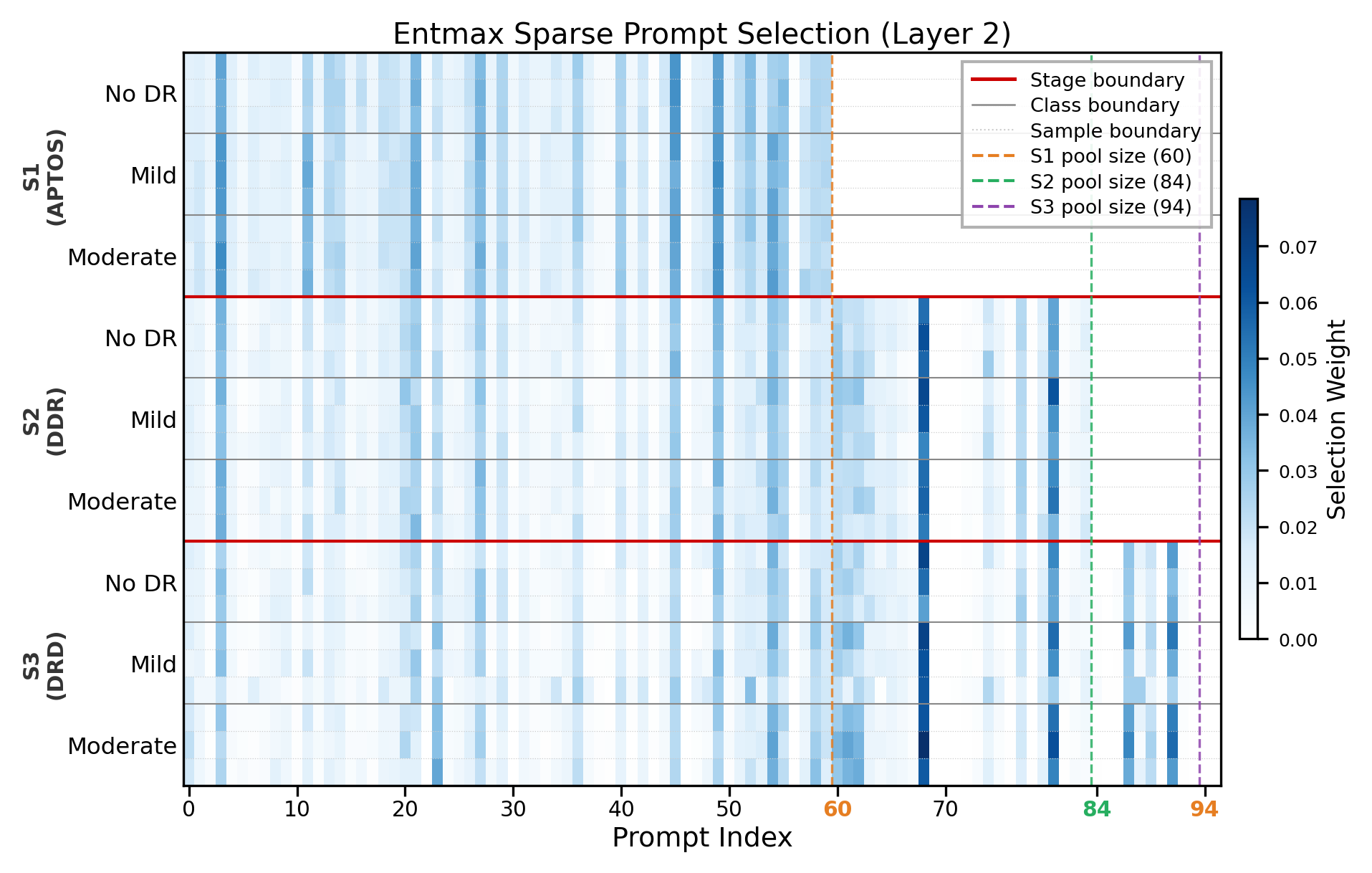}
\caption{Stage-wise prompt selection with PUDD-triggered pool expansion (60$\rightarrow$84$\rightarrow$94) and sparse $\alpha$-entmax routing on the DR dataset. Corresponding results on the other datasets are provided in Sup.~\ref{supp:E}.}
\label{figure3}
\end{figure*}

\begin{figure*}[ht!]
\centering
\includegraphics[width=1.0\textwidth]{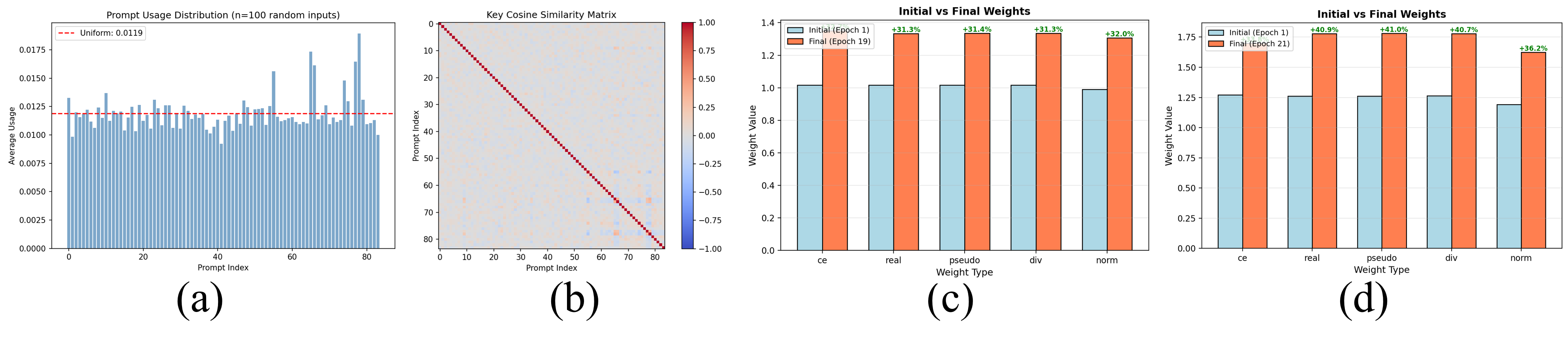}
\caption{Prompt usage and optimization dynamics on the DR dataset: (a) balanced selection, (b) no redundancy after expansion, (c--d) uncertainty-weighted loss rebalancing across stages. Corresponding results for the other datasets are provided in Sup.~\ref{supp:E}.}
\label{figure4}
\end{figure*}

\noindent\textbf{Empirical Analysis of Residual SODAP.}
Figs.~\ref{figure3} and \ref{figure4} present analyses verifying that Residual SODAP operates as intended.
Fig.~\ref{figure3} illustrates how the per-input prompt selection pattern evolves as the stage progresses. We randomly sample three instances per class from each stage for analysis. As the learning stream advances, PUDD detects distributional shifts and the prompt pool size expands incrementally from 60$\rightarrow$84
$\rightarrow$94. The model fine-tunes the selected prompts according to each sample's characteristics for inference, and continues to leverage both existing and newly acquired knowledge even after expansion. Notably, Entmax-based selection concentrates probability mass on a small subset of highly relevant prompts per input, effectively suppressing the unnecessary prompt activation that can arise with softmax-based selection. Results for all layers are provided in Supp.~\ref{supp:D}.
Fig.~\ref{figure4} more directly examines the distributional properties of prompt selection and whether expansion introduces redundancy. 
Fig.~\ref{figure4}(a) visualizes prompt selection frequencies, showing that the excessive reliance on a few specific prompts commonly observed with conventional Top-k selection—is mitigated, yielding a more balanced distribution. Fig.~\ref{figure4}(b) demonstrates that expanding the prompt pool does not produce functionally redundant prompts, suggesting that the model selectively acquires only the necessary information rather than duplicating existing representations.
Figs.~\ref{figure4}(c) and (d) show that the multiple loss terms introduced in Sec.~\ref{subsec:uw} are dynamically adjusted as intended across stages. Stage~1 is excluded from this analysis, as uncertainty weighting is not applied at that stage. Fig.~\ref{figure4}(c) presents the relative magnitudes of the learned loss terms at Stage~2, and Fig.~\ref{figure4}(d) shows the corresponding results at Stage~3. Overall, the contribution of each loss term is automatically rebalanced according to the stage and data characteristics, maintaining a well-composed learning signal without manual tuning.

\begin{table}[t]
\centering
\caption{\textbf{Sequential ablation} on the DR continual learning benchmark (APTOS $\rightarrow$ DDR $\rightarrow$ DRD).
We remove \emph{one} component at a time from the full model.
$\Delta$ reports the absolute change from the full model in \textbf{percentage points} (pp):
$\Delta\mathrm{AvgACC}=100\times(\mathrm{AvgACC}-0.850)$ and
$\Delta\mathrm{AvgF}=100\times(\mathrm{AvgF}-0.047)$
(higher AvgACC and lower AvgF are better).
\cmark{} indicates the component is enabled.}
\label{tab:ablation}
\setlength{\tabcolsep}{3pt}
\small
\resizebox{1.0\linewidth}{!}{%
\begin{tabular}{l|cccccccc|cc|cc}
\toprule
& \multicolumn{8}{c|}{\textbf{Components}} & \multicolumn{2}{c|}{\textbf{Metric}} & \multicolumn{2}{c}{\textbf{vs. Full (pp)}} \\
\cmidrule(lr){2-9}\cmidrule(lr){10-11}\cmidrule(lr){12-13}
\textbf{Method}
& \textbf{SODAP} & \textbf{PUDD} & \textbf{Distill} & \textbf{Pseudo}
& $\boldsymbol{\mathcal{L}_{div}}$ & $\boldsymbol{\mathcal{L}_{norm}}$ & \textbf{UW} & \textbf{QE}
& AvgACC$\uparrow$ & AvgF$\downarrow$
& $\Delta$AvgACC$\uparrow$ & $\Delta$AvgF$\downarrow$ \\
\midrule

\rowcolor{gray!12} \textbf{Full Model}
& \cmark & \cmark & \cmark & \cmark & \cmark & \cmark & \cmark & \cmark
& 0.850$\pm$0.001 & 0.047$\pm$0.003
& +0.0 & +0.0 \\

\midrule
\multicolumn{13}{l}{\textit{(1) Prompting / drift handling}}\\
w/o Query Enhancer (QE)
& \cmark & \cmark & \cmark & \cmark & \cmark & \cmark & \cmark &
& 0.808$\pm$0.013 & 0.032$\pm$0.007
& -4.2 & -1.5 \\

w/o PUDD
& \cmark &  & \cmark & \cmark & \cmark &  & \cmark & \cmark
& 0.841$\pm$0.004 & 0.028$\pm$0.009
& -0.9 & -1.9 \\

w/o Residual SODAP (SODAP)
&  & \cmark & \cmark & \cmark & \cmark & \cmark & \cmark & \cmark
& 0.831$\pm$0.004 & 0.027$\pm$0.003
& -1.9 & -2.0 \\

\midrule
\multicolumn{13}{l}{\textit{(2) Classifier knowledge preservation}}\\
w/o Pseudo Replay
& \cmark & \cmark & \cmark &  & \cmark & \cmark & \cmark & \cmark
& 0.835$\pm$0.008 & 0.028$\pm$0.004
& -1.5 & -1.9 \\

w/o Distillation
& \cmark & \cmark &  &  & \cmark & \cmark & \cmark & \cmark
& 0.828$\pm$0.009 & 0.032$\pm$0.009
& -2.2 & -1.5 \\

\midrule
\multicolumn{13}{l}{\textit{(3) Objective regularization / balancing}}\\
w/o $\mathcal{L}_{div}$
& \cmark & \cmark & \cmark & \cmark &  & \cmark & \cmark & \cmark
& 0.818$\pm$0.011 & 0.072$\pm$0.001
& -3.2 & +2.5 \\

w/o $\mathcal{L}_{norm}$
& \cmark & \cmark & \cmark & \cmark & \cmark &  & \cmark & \cmark
& 0.838$\pm$0.010 & 0.008$\pm$0.001
& -1.2 & -3.9 \\

w/o Uncertainty Weighting (UW)
& \cmark & \cmark & \cmark & \cmark & \cmark & \cmark &  & \cmark
& 0.838$\pm$0.004 & 0.011$\pm$0.005
& -1.2 & -3.6 \\

\bottomrule
\end{tabular}%
}
\end{table}

\noindent\textbf{Ablation Study.}
Table~\ref{tab:ablation} presents a sequential ablation on the DR domain-incremental benchmark (APTOS $\rightarrow$ DDR $\rightarrow$ DRD). We remove one component at a time from the full model and assess the accuracy--forgetting trade-off between AvgACC and AvgF. Among the prompting/drift-handling modules, removing the QE causes the largest drop in AvgACC ($-4.2$ pp), indicating that query refinement is critical for reliable prompt selection throughout the training stream. Routing-related ablations (w/o QE, w/o PUDD, and w/o SODAP) tend to reduce AvgF (i.e., mitigate forgetting), but this comes with non-trivial accuracy loss, suggesting a shift toward a more conservative update regime and a less favorable overall trade-off. For classifier knowledge preservation, both pseudo replay and distillation are important. Removing pseudo replay decreases AvgACC by $1.5$ pp, and removing distillation results in a $2.2$ pp drop, confirming that lightweight preservation mechanisms remain beneficial even under a frozen-backbone prompt-based setting. Finally, objective regularization and balancing terms strongly influence stability and performance. Dropping $\mathcal{L}_{div}$ degrades both dimensions, reducing AvgACC by $3.2$ pp while increasing AvgF by $2.5$ pp, which highlights the role of diversity regularization in preventing collapse and preserving past knowledge. In contrast, removing $\mathcal{L}_{\mathrm{norm}}$ or UW leads to a modest drop in AvgACC ($-1.2$ pp) while generally reducing AvgF, highlighting an inherent trade-off between maximizing accuracy and mitigating forgetting. In other words, some variants reduce forgetting at the expense of accuracy, corresponding to different operating points on the trade-off curve, whereas our full model targets a balanced regime that maintains high accuracy without excessively worsening forgetting. Overall, the full model provides the most favorable trade-off among the tested variants (AvgACC $0.850$, AvgF $0.047$).

\section{Conclusion}
\label{sec:conclusion}
This study proposes \textbf{Residual SODAP}, a rehearsal-free domain-incremental learning framework that simultaneously achieves representation adaptation and classifier-level knowledge preservation. By integrating $\alpha$-entmax-based sparse prompt selection with residual fusion, \textbf{Statistical Knowledge Preservation} via statistics-based pseudo-feature replay to alleviate classifier-level forgetting, \textbf{PUDD}, and \textbf{uncertainty weighting}, the proposed method effectively mitigates catastrophic forgetting without relying on task identity or access to past data. Extensive experiments on two medical imaging benchmarks and a general-domain dataset demonstrate consistent improvements over prior methods across different continual learning paradigms.
Furthermore, ablation studies and qualitative analyses validate the contribution of each component and confirm that the overall framework behaves in accordance with its design intent.
As future work, we will extend Residual SODAP to class-incremental scenarios with evolving label spaces and develop more principled drift-aware prompt management, including adaptive expansion schedules and capacity control under compute and memory constraints. We also plan to refine classifier-level preservation by designing class-conditional statistical pseudo replay and by improving the calibration of uncertainty weighting to better handle long-tailed, noisy medical labels and subtle domain shifts.

\label{sec:Conclusion}
 
\clearpage  


%
%
\bibliographystyle{splncs04}
\bibliography{main}

\vfill\pagebreak
\appendix

\section*{Supplementary Material}

\section*{Supplementary Material Overview}

This supplementary material is organized as follows.
\begin{itemize}
  \item \textbf{Sec.~\ref{supp:A} Implementation and Reproducibility Details.}
  We describe the rehearsal-free domain-incremental learning (DIL) setting, dataset construction and preprocessing for DR (APTOS$\rightarrow$DDR$\rightarrow$DRD), Skin Cancer (ISIC$\rightarrow$HAM$\rightarrow$DERM7), and CORe50 (11-stage stream), and report full training configurations and standardized baseline hyperparameters (Sup.~Tables~\ref{Table2}--\ref{tab:core50_stats}, Tables~\ref{maine_setting}--\ref{maine_setting_2}).

  \item \textbf{Sec.~\ref{supp:B} Additional Analysis: Classifier Instability under Domain Shifts.}
  We present a diagnostic cross-composition analysis (Backbone $\times$ Classifier) following \cite{liu2020more} under the Skin Cancer DIL protocol, and report full $3\times 3$ cross-stage compositions for the baseline \cite{kim2024one} (Sup.~Fig.~\ref{figure1_2}) and our method (Sup.~Fig.~\ref{figure1_3}).

  \item \textbf{Sec.~\ref{supp:C} Continual Learning Metrics.}
  We provide formal definitions and interpretations of the evaluation metrics used throughout the paper, including AvgACC and AvgF.

  \item \textbf{Sec.~\ref{supp:D} Additional Analysis: Entmax Sparse Prompt Selection Across All Layers.}
  We analyze $\alpha$-entmax sparse prompt selection across stages and layers, including stage-wise expansion, selection frequency, and prompt similarity statistics (Sup.~Fig.~\ref{figure5}).

  \item \textbf{Sec.~\ref{supp:E} Additional Results on Other Datasets.}
  We provide qualitative results on Skin Cancer and CORe50, including stage-wise expansion, balanced prompt usage, and uncertainty-weighted loss dynamics (Sup.~Figs.~\ref{figure6}--\ref{figure8}).

  \item \textbf{Sec.~\ref{supp:F} Use of AI Tools Disclosure.}
  We disclose limited auxiliary use of AI tools for writing support (ChatGPT) and code maintenance (Cursor/Claude CLI); all scientific content and final decisions were made by the authors.
\end{itemize}


\section{Implementation and Reproducibility Details}
\label{supp:A}

\noindent\textbf{Domain Incremental Continual Learning Setting}
Following \cite{wang2024comprehensive}, we consider a rehearsal-free DIL setting where original data from past domains cannot be stored. Domains arrive sequentially at each stage $t\in\{1,\dots,T\}$, and only the current dataset $\mathcal{D}_t=\{(x_i,y_i)\}_{i=1}^{n_t}$ is accessible, where $x_i$ and $y_i$ denote the image and its label, respectively. Past data $\mathcal{D}_{<t}$ is unavailable, and no domain-switch signal or task identity is provided at inference time. This setting realistically reflects the domain shifts and domain uncertainty frequently encountered in practical deployment scenarios. The goal is to train a single model that effectively adapts to sequentially arriving domains while minimizing performance degradation (forgetting) on previously seen domains.

\begin{table}[ht]
\centering
\caption{Dataset statistics for the DR benchmarks after remapping to three classes (Normal, NPDR, PDR).}
\label{Table2}
\begin{tabular}{lrrr}
\hline
\textbf{Dataset} & \textbf{Normal (0)} & \textbf{NPDR (1)} & \textbf{PDR (2)} \\
\hline
APTOS~\cite{aptos2019-blindness-detection} & 1805 & 1562 & 295 \\
DDR~\cite{LI2019}                         & 6266 & 5343 & 913 \\
DRD~\cite{diabetic-retinopathy-detection} & 5000 & 8608 & 708 \\
\hline
\end{tabular}
\end{table}

\noindent\textbf{Primary benchmark (Diabetic Retinopathy).}
Our main experiments use three publicly available diabetic retinopathy (DR) datasets: APTOS, DDR, and DRD.
Since some benchmarks do not provide ground-truth labels for their official test sets, we follow the preprocessing protocol of \cite{kobat2022automated} to construct consistent train/validation/test splits.
Following the same study, we remap the original five-level DR grading (normal, mild, moderate, severe, proliferative) into three clinically meaningful classes (Normal, NPDR, PDR): mild, moderate, and severe are merged into NPDR, while normal and PDR are kept as-is.
The resulting data distribution and dataset sources are summarized in Sup.~Table~\ref{Table2}.
Under the domain-incremental continual learning setting, the training stream is denoted as:
\[
D=\{D_1,D_2,\ldots,D_n\},\qquad D_n=\{X_{n,b},Y_{n,b}\},
\]
where $n$ is the stage index, $X_{n,b}$ and $Y_{n,b}$ are the input image and its label, respectively, and $b$ is the sample index within a batch of size~$B$.

\begin{table}[ht]
\centering
\caption{Distribution of Skin Cancer datasets used for training}
\label{Table4}
\begin{tabular}{lrr}
\hline
\textbf{Dataset} & \textbf{Benign} & \textbf{Malignant} \\
\hline
ISIC~\cite{rotemberg2021patient} & 584 & 584 \\
HAM~\cite{codella2019skin}       & 1113 & 1113 \\
DERM7~\cite{kawahara2018seven}   & 252 & 252 \\
\hline
\end{tabular}
\end{table}

\noindent\textbf{Supplementary external benchmark (Skin Cancer).}
To evaluate generalization beyond DR, we conduct an additional experiment under the same domain-incremental protocol using three skin cancer datasets in the order ISIC $\rightarrow$ HAM $\rightarrow$ DERM7, forming three sequential stages.
Following \cite{cassidy2022analysis} on clinically relevant lesion grouping in ISIC, we remap the original diagnostic categories into a binary setting: lesions requiring urgent clinical intervention are labeled as malignant, and the rest as benign.
Dataset statistics are provided in Sup.~Table~\ref{Table4}.
All datasets are split into train/validation/test at an 8:1:1 ratio.
All other settings, including the model, training schedule, hyperparameters, and evaluation metrics, remain identical to the DR experiments; only the datasets are changed. Due to the smaller dataset scale, the batch size is set to 16.

\noindent\textbf{Supplementary external benchmark (CORe50).}
To demonstrate that the proposed method is not limited to medical data and generalizes to standard DIL settings, we additionally evaluate on the public CORe50 dataset, which is widely used for DIL performance assessment. CORe50 consists of 10 classes and includes a total of 11 domain shifts induced by variations in acquisition conditions such as background and illumination. Accordingly, we construct an 11-stage domain-incremental learning stream and measure performance as the stages progress. The dataset split is summarized in Table~\ref{tab:core50_stats}, where the data are partitioned into 70/15/15 for train/validation/test. These results demonstrate that our model is not restricted to the medical domain and remains robust even under a large number of domain shifts.

\begin{table}[ht]
\centering
\caption{Per-class training sample counts for each session in CORe50~\cite{lomonaco2017core50}. Each session serves as one stage in our 11-stage domain-incremental continual learning setup (70/15/15 split).}
\label{tab:core50_stats}
\resizebox{\columnwidth}{!}{%
\begin{tabular}{clrrrrrrrrrrr}
\hline
\textbf{Class} & \textbf{Category} & \textbf{Stage 1} & \textbf{Stage 2} & \textbf{Stage 3} & \textbf{Stage 4} & \textbf{Stage 5} & \textbf{Stage 6} & \textbf{Stage 7} & \textbf{Stage 8} & \textbf{Stage 9} & \textbf{Stage 10} & \textbf{Stage 11} \\
\hline
0 & Plug adapters   & 1,064 & 1,066 & 1,066 & 1,064 & 1,066 & 1,064 & 1,066 & 1,064 & 1,066 & 1,066 & 1,066 \\
1 & Mobile phones   & 1,048 & 1,052 & 1,052 & 1,052 & 1,052 & 1,051 & 1,049 & 1,050 & 1,052 & 1,052 & 1,047 \\
2 & Scissors        & 1,070 & 1,067 & 1,067 & 1,069 & 1,064 & 1,070 & 1,070 & 1,066 & 1,067 & 1,067 & 1,070 \\
3 & Light bulbs     & 1,041 & 1,042 & 1,040 & 1,039 & 1,033 & 1,039 & 1,039 & 1,040 & 1,043 & 1,041 & 1,041 \\
4 & Cans            & 1,039 & 1,038 & 1,040 & 1,041 & 1,041 & 1,038 & 1,040 & 1,040 & 1,039 & 1,039 & 1,038 \\
5 & Glasses         & 1,064 & 1,059 & 1,065 & 1,065 & 1,058 & 1,064 & 1,065 & 1,062 & 1,066 & 1,065 & 1,064 \\
6 & Balls           & 1,047 & 1,046 & 1,045 & 1,045 & 1,049 & 1,048 & 1,045 & 1,047 & 1,046 & 1,044 & 1,048 \\
7 & Markers         & 1,002 & 1,009 & 1,002 & 1,011 & 1,014 & 1,002 & 1,020 & 1,001 & 1,017 & 1,007 & 1,004 \\
8 & Cups            & 1,060 & 1,057 & 1,060 & 1,062 & 1,052 & 1,059 & 1,054 & 1,060 & 1,053 & 1,057 & 1,072 \\
9 & Remote controls & 1,057 & 1,054 & 1,057 & 1,048 & 1,047 & 1,057 & 1,047 & 1,058 & 1,046 & 1,052 & 1,043 \\
\hline
\multicolumn{2}{l}{\textbf{Total}} & \textbf{10,492} & \textbf{10,490} & \textbf{10,494} & \textbf{10,496} & \textbf{10,476} & \textbf{10,492} & \textbf{10,495} & \textbf{10,488} & \textbf{10,495} & \textbf{10,490} & \textbf{10,493} \\
\hline
\end{tabular}%
}
\end{table}

\noindent\textbf{Training hyperparameters Setting.}
The experimental protocol follows the configurations summarized in Table~\ref{maine_setting} and Table~\ref{maine_setting_2}. We adhere as closely as possible to the base setup described in Table~\ref{maine_setting}, while the detailed hyperparameters for the compared continual learning (CL) methods are standardized according to Table~\ref{maine_setting_2}. The CL models used in our experiments are implemented based on well-established libraries~\cite{boschini2022class,buzzega2020dark}. When certain methods were not directly available or required adaptation to match our experimental setting, they were carefully re-implemented under the same evaluation protocol.

\begin{table}[t]
\centering
\caption{Experimental Settings.}
\label{maine_setting}
\setlength{\tabcolsep}{6pt}
\renewcommand{\arraystretch}{1.15}
\begin{tabular}{@{}l l@{}}
\toprule
\textbf{Category} & \textbf{Setting} \\
\midrule
\multicolumn{2}{@{}l}{\textbf{Hardware}} \\
GPU & 2$\times$ NVIDIA V100 \\
RAM & 256\,GB \\
\addlinespace
\multicolumn{2}{@{}l}{\textbf{Optimization}} \\
Optimizer & AdamW \\
Learning rate & 1e{-3} \\
Scheduler & Cosine \\
Batch size & 64 \\
Epochs & 100 \\
Early-stop patience & 5 \\
\addlinespace
\multicolumn{2}{@{}l}{\textbf{Model / Method}} \\
Prompt pool size ($N$) & 60 \\
Bottleneck dimension ($d_a$) & 512 \\
Memory slot size ($M$) & 9 \\
Drift threshold ($\theta$) & 0.7 \\
Feature dimension ($D$) & 768 \\
\addlinespace
\multicolumn{2}{@{}l}{\textbf{Input}} \\
Image size & 224$\times$224 \\
\bottomrule
\end{tabular}
\end{table}

\begin{table*}[t]
\centering
\caption{Key hyperparameters for continual learning baselines.}
\label{maine_setting_2}
\setlength{\tabcolsep}{6pt}
\renewcommand{\arraystretch}{1.1}
\begin{tabular}{@{}l l c@{}}
\toprule
\textbf{Method} & \textbf{Hyperparameter} & \textbf{Value} \\
\midrule
\multirow{3}{*}{CODA-Prompt} 
& Pool size (\texttt{coda\_pool\_size}) & 100 \\
& Prompt length (\texttt{coda\_prompt\_length}) & 4 \\
& Ortho. reg. (\texttt{coda\_ortho\_mu}) & 0.1 \\
\midrule
\multirow{7}{*}{DualPrompt} 
& Prompt length (\texttt{dual\_prompt\_length}) & 4 \\
& Pool size (\texttt{dual\_pool\_size}) & 100 \\
& Top-$k$ (\texttt{dual\_top\_k}) & 5 \\
& G-prompt length (\texttt{dual\_g\_prompt\_length}) & 4 \\
& G layers (\texttt{dual\_g\_prompt\_layers}) & [0, 1] \\
& E layers (\texttt{dual\_e\_prompt\_layers}) & [2, 3, 4] \\
& Head type (\texttt{dual\_head\_type}) & \texttt{token} \\
\midrule
\multirow{3}{*}{DER++} 
& $\alpha$ (preserve past) & 0.5 \\
& $\beta$ (relearn past) & 0.5 \\
& Buffer size & 30 or 64 (CORe50) \\
\midrule
\multirow{3}{*}{Online EWC} 
& Momentum (\texttt{optim\_mom}) & 0.9 \\
& Weight decay (\texttt{optim\_wd}) & 5e{-4} \\
\midrule
\multirow{3}{*}{MoE Adapters} 
& Prompt template & \texttt{a photo of \{\}.} \\
& CLIP backbone & \texttt{ViT-B/16} \\
& Virtual BS iters. (\texttt{virtual\_bs\_n}) & 1 \\
\midrule
\multirow{2}{*}{OS (++) Prompt} 
& Prompt Dim  & 768 \\
& Pool size & 150 \\
\bottomrule
\end{tabular}
\end{table*}

\section{Additional Analysis: Classifier Instability under Domain Shifts}
\label{supp:B}
\noindent\textbf{Motivation.}
Prompt-based continual learning (PCL) can effectively adapt representations under sequential domain shifts, but it remains unclear whether prompt improvements alone are sufficient to prevent forgetting throughout the entire learning process.
Following the diagnostic perspective of \cite{liu2020more}, we disentangle the contributions of backbone representations and classifier decision boundaries to forgetting in domain-incremental learning (DIL).

\noindent\textbf{Experimental scenario.}
We construct a 3-stage DIL skin lesion diagnosis scenario where the class space is fixed and only the input domain changes sequentially:
ISIC~\cite{rotemberg2021patient} $\rightarrow$ HAM~\cite{codella2019skin} $\rightarrow$ DERM7~\cite{kawahara2018seven}.
A recent PCL method~\cite{kim2024one} is trained under this protocol. Detailed dataset descriptions are provided in Sup.~\ref{supp:A}.

\noindent\textbf{Cross-composition protocol (Backbone $\times$ Classifier).}
We decompose the model trained at stage $t\in\{1,2,3\}$ into its backbone $B_t$ and classifier head $C_t$.
After training each stage, we store the corresponding $(B_t, C_t)$ pair and evaluate all cross-combinations $\{(B_i, C_j)\}_{i,j=1}^{3}$, yielding $3\times 3$ composite models.
This allows us to attribute performance degradation to
(i)~representation-level factors (varying $B_i$ with fixed $C_j$) or
(ii)~classifier-level factors (varying $C_j$ with fixed $B_i$).

\noindent\textbf{Key observation.}
Prompt-based adaptation generally benefits backbone representations, tending to maintain or improve performance when paired with an appropriate classifier.
However, as DIL progresses, degradation at the classifier level becomes increasingly pronounced.
Specifically, the performance drop caused by swapping the classifier head is larger and more systematic than that caused by swapping the backbone, suggesting that decision-boundary instability in the classifier can be a dominant source of forgetting under domain shifts.
This trend is consistent with the analysis of \cite{liu2020more}, which posits that decision-boundary collapse primarily occurs at the classifier level under sequential learning, leading to drops in both accuracy and F1-score.

\noindent\textbf{Implication.}
These results indicate that improvements to prompt structures and pool designs are necessary but not sufficient.
Robust continual learning under domain shifts requires explicitly addressing classifier-level knowledge preservation alongside prompt-based representation adaptation, which provides the core motivation for Residual SODAP that integrates preservation mechanisms at both levels.

\begin{figure*}[ht!]
\centering
\includegraphics[width=1.0\textwidth]{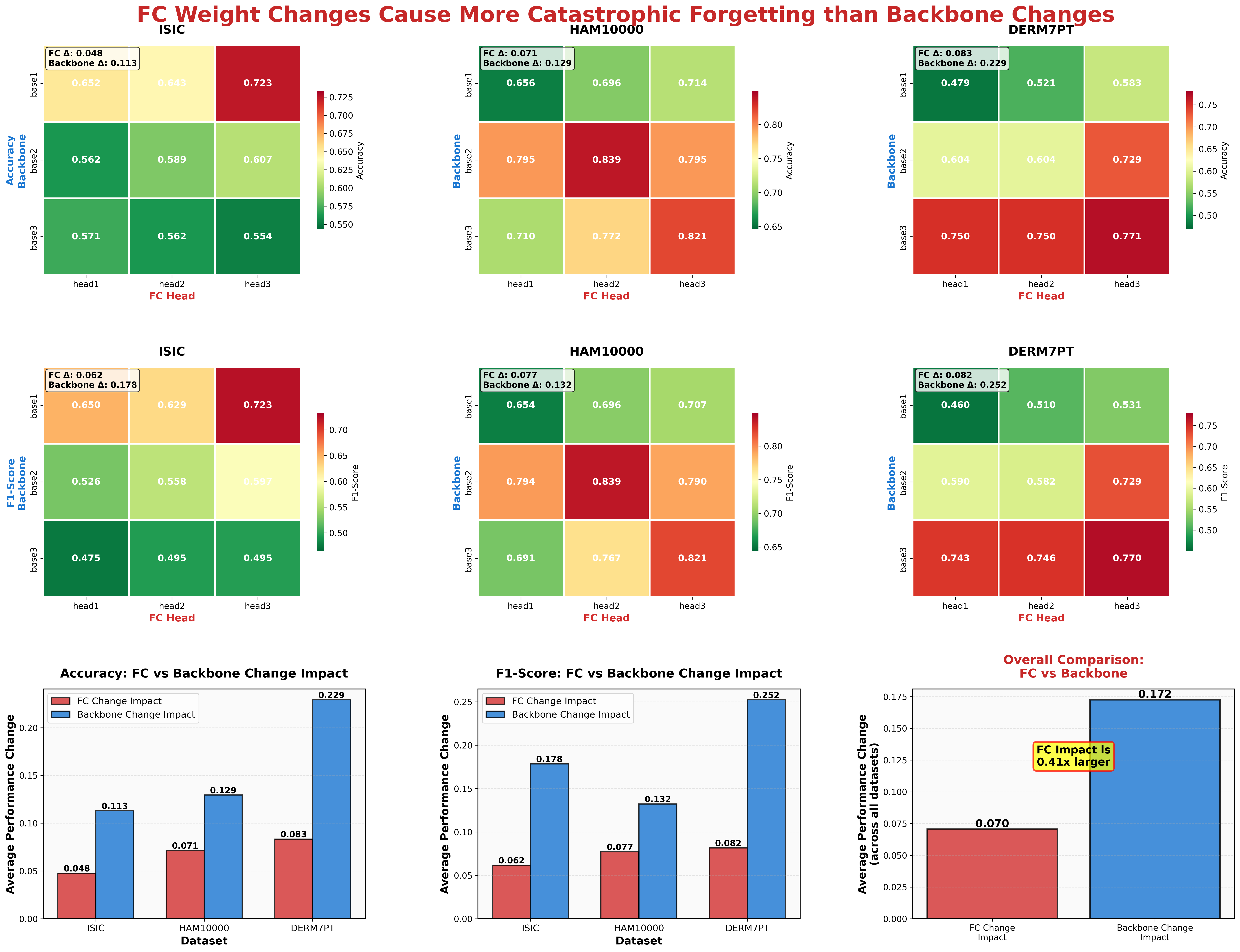}
\caption{Results of cross-stage backbone--classifier compositions ($3\times 3$) under the skin-cancer DIL protocol. The baseline visualization follows the prompt-based continual learning method of \cite{kim2024one}.}
\label{figure1_2}
\end{figure*}

\begin{figure*}[ht!]
\centering
\includegraphics[width=1.0\textwidth]{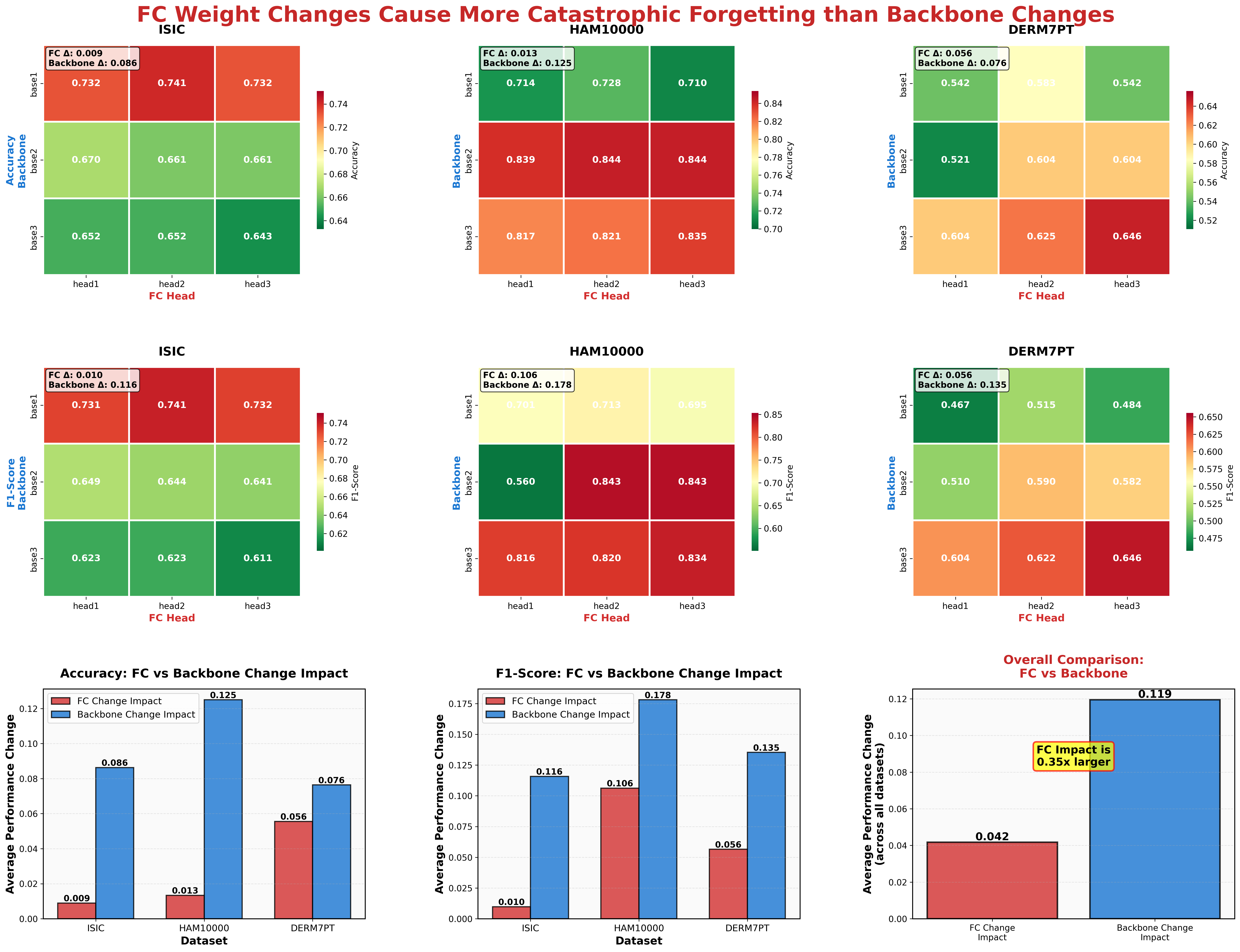}
\caption{Results of cross-stage backbone--classifier compositions ($3\times 3$) under the skin-cancer DIL protocol. This figure visualizes \textbf{our method} using the same cross-composition protocol; the corresponding baseline visualization based on \cite{kim2024one} is provided in Sup.~Fig.~\ref{figure1_2}.}
\label{figure1_3}
\end{figure*}

\noindent\textbf{Additional results.}
Fig.~\ref{figure1} provides a summary of this analysis, and Sup.~Fig.~\ref{figure1_2} reports per-stage results and the full $3\times 3$ cross-composition outcomes.
\textbf{Moreover, Sup.~Fig.~\ref{figure1_3} visualizes our method under the same cross-composition protocol as Sup.~Fig.~\ref{figure1_2}, where the classifier-induced degradation is noticeably mitigated, indicating improved stability of decision boundaries over the domain-incremental stream.}

\section{Continual Learning Metrics}
\label{supp:C}
\begin{equation}
\mathrm{AvgACC}
= \frac{1}{T}\sum_{i=1}^{T}\left(\frac{1}{i}\sum_{j=1}^{i} R_{i,j}\right).
\end{equation}

\noindent\textbf{Symbols.}
$T$ denotes the total number of stages (or tasks), and $R_{i,j}$ is the accuracy on task $j$ measured right after training up to stage $i$.

\noindent\textbf{How it is computed.}
At each stage $i$, we first average the accuracies over all tasks observed so far (from $1$ to $i$).
We then average these stage-wise values over $i=1,\dots,T$ to obtain $\mathrm{AvgACC}$.

\noindent\textbf{Interpretation.}
$\mathrm{AvgACC}$ summarizes the overall performance profile across the entire learning trajectory, rather than only the final outcome.
It therefore reflects how consistently the model maintains performance as new stages arrive.

\begin{equation}
\mathrm{AvgF}
= \frac{1}{T-1}\sum_{j=1}^{T-1}\left(\max_{i \in \{j,\dots,T\}} R_{i,j} - R_{T,j}\right).
\end{equation}

\noindent\textbf{Symbols.}
$T$ is the total number of stages (or tasks), and $R_{i,j}$ is the accuracy on task $j$ after training up to stage $i$.
$R_{T,j}$ denotes the final accuracy on task $j$ after completing all $T$ stages, while $\max_{i\in\{j,\dots,T\}}R_{i,j}$ is the best accuracy achieved on task $j$ at any point during training.

\noindent\textbf{How it is computed.}
For each earlier task $j$, we compute the drop from its peak performance during training to its final performance after stage $T$.
$\mathrm{AvgF}$ is the average of this drop over $j=1,\dots,T-1$.

\noindent\textbf{Interpretation.}
$\mathrm{AvgF}$ directly quantifies forgetting by comparing a task's best historical performance with its end-of-training performance.
Larger values indicate more severe forgetting, while values near zero imply strong retention.

\section{Additional Analysis: Entmax Sparse Prompt Selection Across All Layers}
\label{supp:D}
\begin{figure*}[ht!]
\centering
\includegraphics[width=1.0\textwidth]{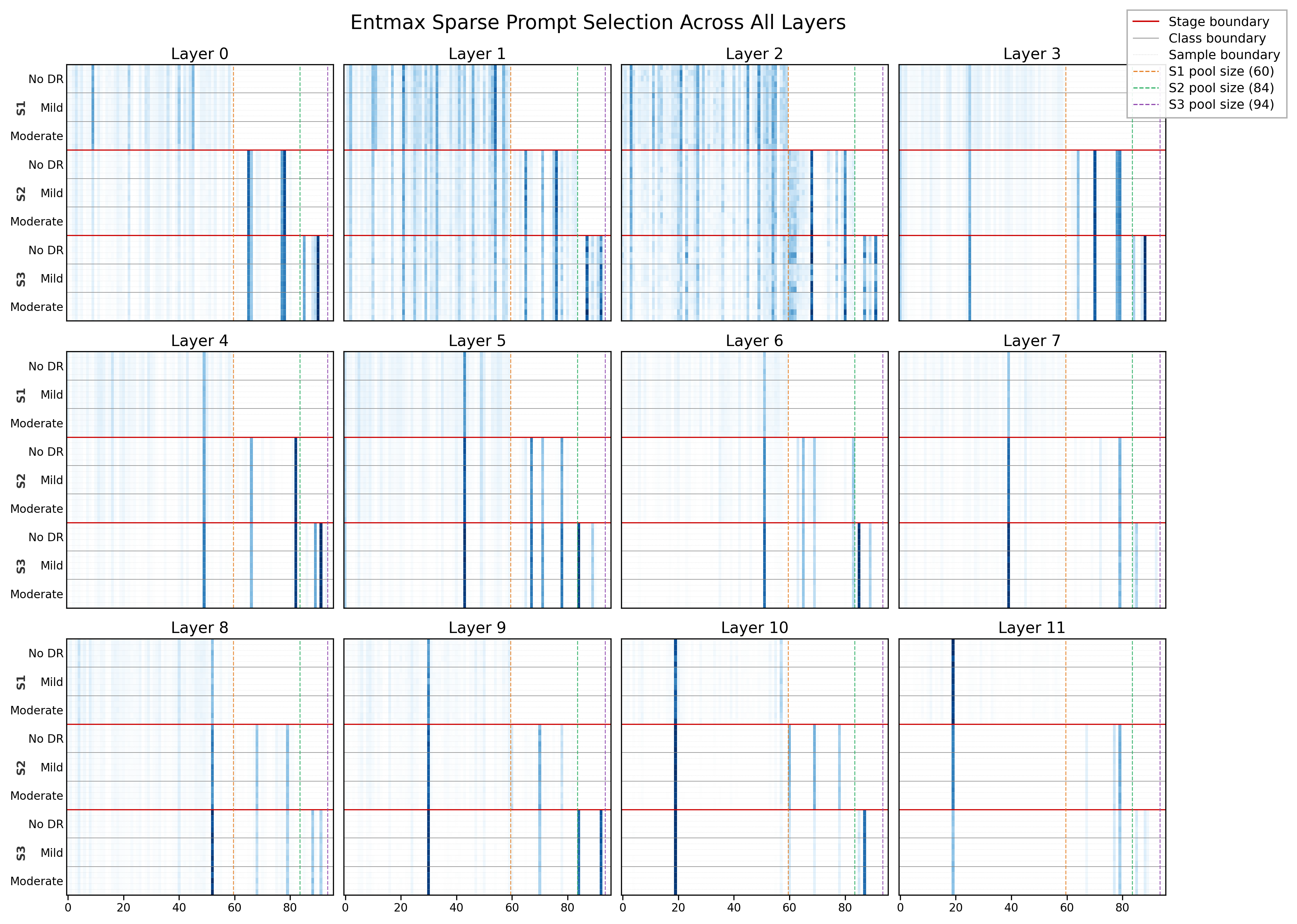}
\caption{Residual SODAP prompt pool analysis: (a) stage-wise prompt count ($60 \rightarrow 84 \rightarrow 94$), (b) prompt selection ratio on 100 random samples, and (c) prompt cosine similarity.}
\label{figure5}
\end{figure*}

As illustrated in Fig.~\ref{figure5}, this experiment qualitatively analyzes how the proposed Residual SODAP selects prompts from the entire prompt pool via $\alpha$-entmax as the pool expands, and how these selection patterns vary across stages and layers. For each stage, we load the corresponding trained weights and randomly sample three instances per class for analysis.

The results show that, even after prompt pool expansion, previously learned prompts and newly introduced prompts are selected in a balanced manner, consistent with our design intention. In other words, while expanded prompts are utilized for adapting to new domains, prompts learned from earlier domains continue to be selected, indicating preservation of prior knowledge. We further observe that prompt selection behavior differs across layers. In early layers (0--2), prompt selection tends to be relatively balanced across multiple prompts. In intermediate layers (3--9), a subset of key prompts receives higher probability mass and plays a more dominant role in shaping the representation. In later layers, a small number of prompts more strongly transform the feature representation, while the remaining prompts contribute residual refinements.

These observations are consistent with the general characteristic that deeper layers progressively compress and abstract information. They suggest that the selection mechanism of Residual SODAP adaptively leverages prompts according to the representational role of each layer. Overall, $\alpha$-entmax-based sparse selection suppresses unnecessary prompt activation while enabling both existing and newly expanded prompts to be jointly utilized, thereby supporting stable adaptation and knowledge preservation across stages.

\section{Additional Results on Other Datasets.}
\label{supp:E}
\begin{figure*}[ht!]
\centering
\includegraphics[width=1.0\textwidth]{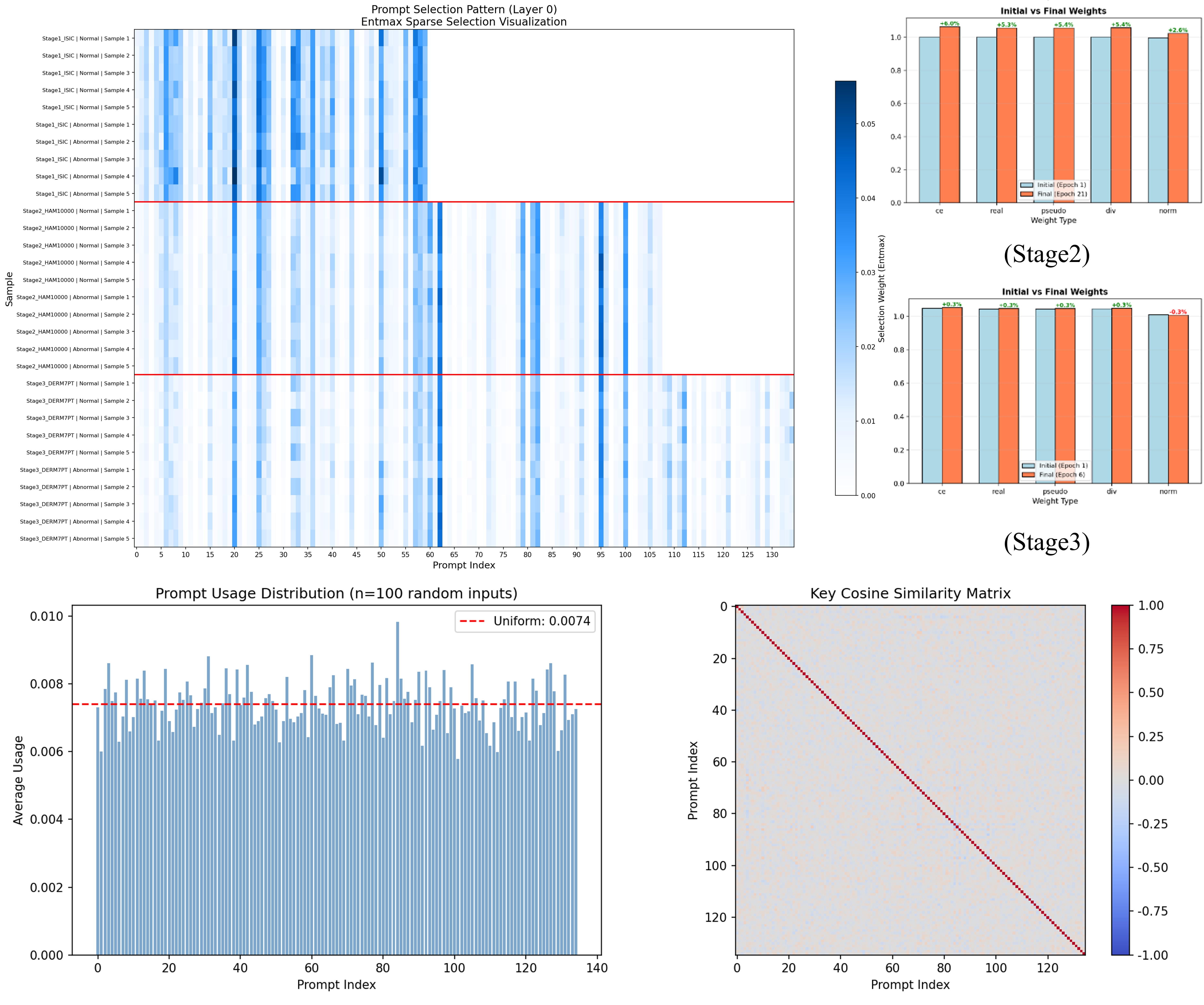}
\caption{Supplementary analysis on the skin-cancer benchmark.}
\label{figure6}
\end{figure*}

\noindent\textbf{Skin cancer.}
As shown in Fig.~\ref{figure6}, the prompt pool expansion behavior and the $\alpha$-entmax-based sparse selection patterns observed on the DR benchmark consistently appear on the skin-cancer benchmark as well. Moreover, the uncertainty-weighting scheme learns diverse loss weights in a stable manner, exhibiting not only monotonic increases but also meaningful decreases when appropriate. We further observe balanced prompt utilization across the pool, and the expanded prompts acquire new information without redundancy, indicating that the additional capacity captures novel domain-specific cues rather than duplicating existing knowledge.

\begin{figure*}[ht!]
\centering
\includegraphics[width=1.0\textwidth]{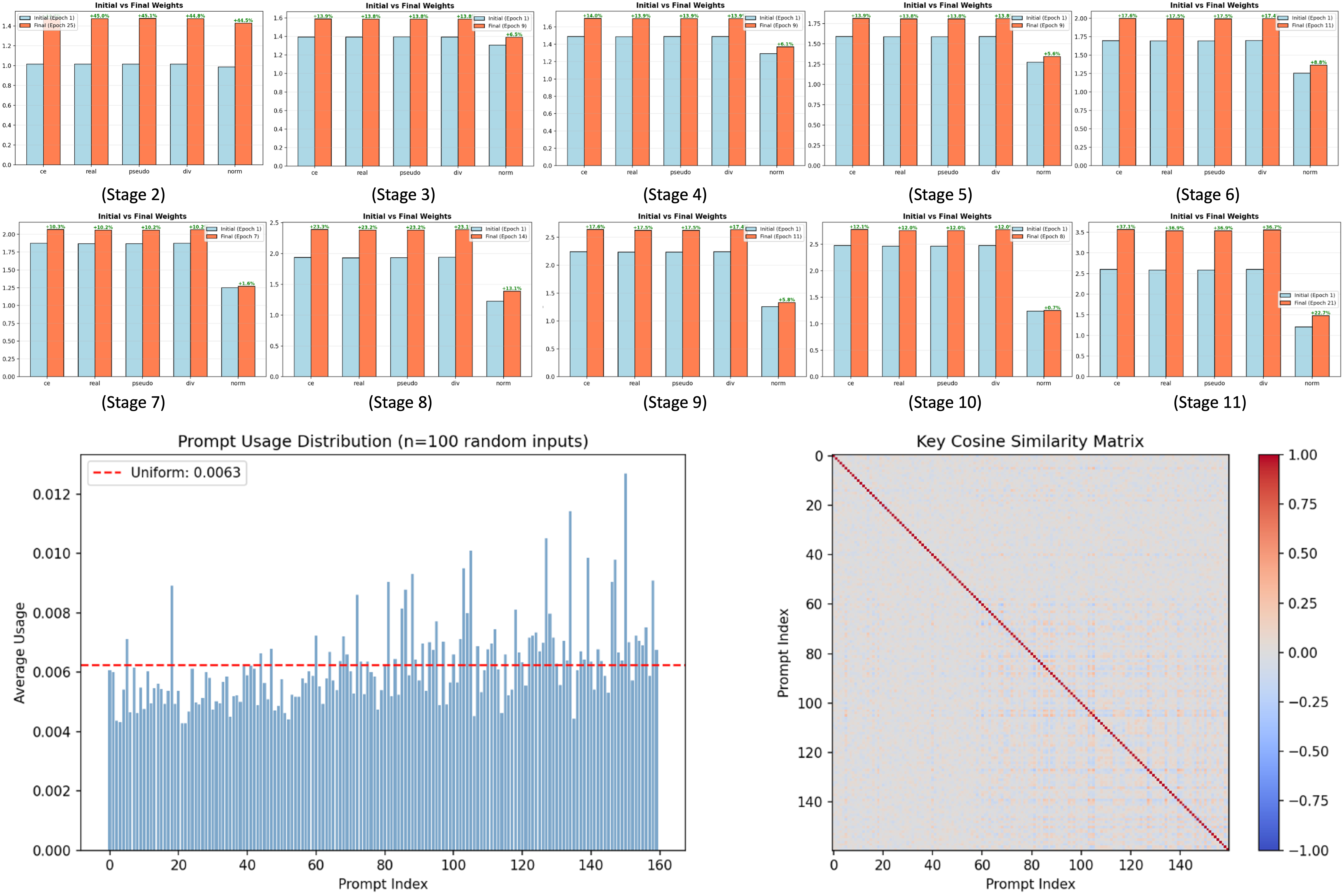}
\caption{Supplementary analysis on the CORe50 benchmark.}
\label{figure7}
\end{figure*}

\begin{figure*}[ht!]
\centering
\includegraphics[width=1.0\textwidth]{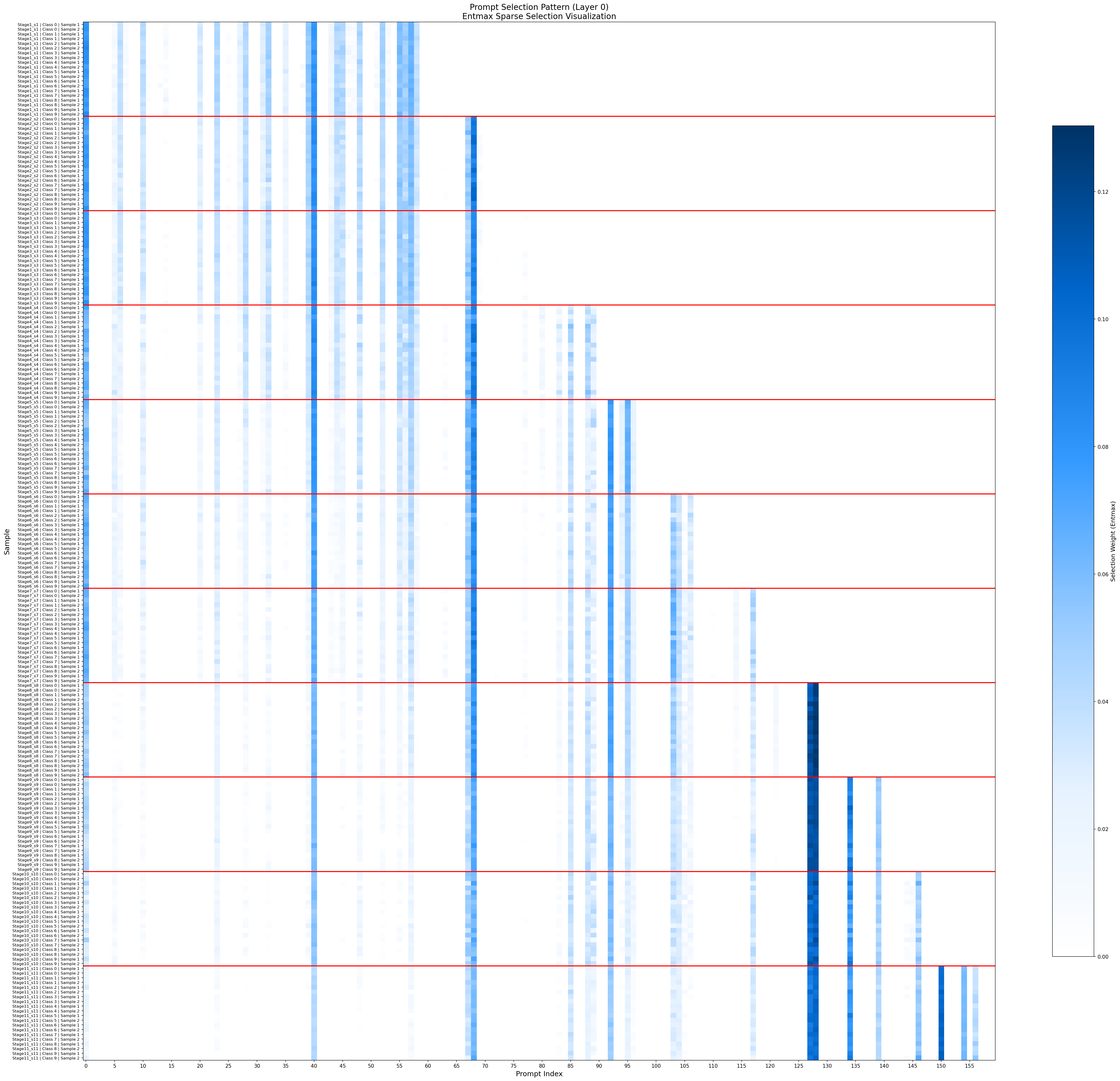}
\caption{Supplementary analysis on the CORe50 benchmark.}
\label{figure8}
\end{figure*}

\noindent\textbf{CORe50.}
As shown in Figs.~\ref{figure7} and \ref{figure8}, the proposed prompt expansion and prompt learning behaviors, together with uncertainty-weighted optimization, operate as intended on the general-domain benchmark. Notably, these patterns remain stable even with the larger number of stages in CORe50, indicating that the framework generalizes well to longer learning streams. Overall, these results provide evidence that our approach is not limited to medical imaging, but can also learn robustly on a general dataset, supporting the generalizability of the proposed components beyond the medical domain.

\section{Use of AI Tools Disclosure.}
\label{supp:F}
We disclose that AI tools were used in a limited and auxiliary capacity during the preparation of this manuscript.
All scientific content---including the research conception, experimental design, result interpretation, core claims, and final writing decisions---was produced by the authors.

ChatGPT was used only for writing support, such as proofreading, 

typo/grammar correction, translation into English, and language polishing.
Cursor and the Claude CLI were used only for code maintenance tasks, including refactoring and cleanup to improve readability (e.g., formatting, comment organization, and minor restructuring).
These tools were not used to generate new experimental results, data, numerical findings, or conclusions.
All final manuscript text and code changes were reviewed and approved by the authors, who take full responsibility for the work.

\end{document}